%% file: main.tex
\documentclass{article}

\usepackage[verbose=true,letterpaper]{geometry}
  \newgeometry{
    textheight=9in,
    textwidth=5.5in,
    top=1in,
    headheight=12pt,
    headsep=25pt,
    footskip=30pt
  }

\usepackage{parskip}

\PassOptionsToPackage{numbers}{natbib}
\usepackage{natbib}

\usepackage{hyperref}
\usepackage[utf8]{inputenc} 
\usepackage[T1]{fontenc}    
\usepackage{url}            
\usepackage{booktabs}       
\usepackage{amsmath}        
\usepackage{amsfonts}       
\usepackage{nicefrac}       
\usepackage{microtype}      
\usepackage{xcolor}         
\usepackage{graphicx}       
\usepackage{amssymb}
\usepackage{wrapfig}        
\graphicspath{{figures/}}

\hypersetup{
    colorlinks=true,
    linkcolor=blue,
    filecolor=magenta,
    urlcolor=magenta,
    citecolor=blue,
    pdftitle={zPocket: Better Pocket Finding with Frozen Co-Folding Features},
}

\input{figures/tables}

\newcommand{\method}{zPocket}
\newcommand{\auprc}{AUPRC}
\newcommand{\qcov}{\texttt{max\_qcov}}
\newif\ifarxiv 
\arxivtrue 

\newif\ificlrfinal
\iclrfinaltrue
\let\zpOrigBibliographystyle\bibliographystyle
\renewcommand{\bibliographystyle}[1]{\zpOrigBibliographystyle{plainnat}}

\date{}

\title{\method{}: Better Pocket Finding with Frozen \\ Co-Folding Features}

\author{
\normalsize \textbf{Tally Portnoi}$^1$ \quad \textbf{Allison M Keys}$^2$ \quad
\textbf{Benjamin R DiFrancesco}$^2$ \quad \textbf{Nate Gruver}$^2$ \\
\normalsize $^1$MIT\quad$^2$Genesis Molecular AI
}

\begin{document}

\maketitle

\begin{abstract}
Identifying protein-ligand binding sites (pockets) is important for understanding protein function and developing drugs. However, no public method has leveraged co-folding models for novel pocket prediction; furthermore, this task lacks a suitable benchmark as existing benchmarks are within the training distribution of co-folding models. We therefore developed \textbf{PocketWatch}, a curated dataset of biologically relevant ligand pockets. The PocketWatch test set consists of unseen pockets stratified by sequence similarity, structural similarity, and crypticity to enable evaluation of generalization to novel and challenging pockets. Using this dataset, we develop \textbf{zPocket}, a method that trains pocket-prediction heads on top of frozen features from co-folding models. We show that zPocket heads outperform baseline methods across all stratifications. Our best zPocket model achieves overall per-target AUPRC of \textbf{0.745} compared to \textbf{0.598} and \textbf{0.595} for the strongest baselines, VN-EGNN and UniSite-1D retrained on the PocketWatch split (\textbf{0.559} and \textbf{0.531} when using the published weights); the widely used P2Rank reaches \textbf{0.477}. We further show that zPocket can be used to outperform baseline methods on top-$N{+}2$ recall of predicted pocket centroids through clustering per-residue predictions or using zPocket to rescore centroids proposed by other models. Finally, we explore two applications of zPocket: crystallographic fragment screens and a proteome-scale scan for uncharacterized pockets on human proteins. Together these results suggest that zPocket heads built on co-folding trunk features can greatly advance ligand binding site prediction.
\end{abstract}

\ifarxiv
\begin{center}
\small
Code: \url{https://github.com/genesistherapeutics/zpocket} \\
Model weights: \url{https://huggingface.co/genesisml/zPocket} \\
Dataset: \url{https://huggingface.co/datasets/genesisml/PocketWatch}
\end{center}
\fi

\section{Introduction}
\label{sec:intro}

Identifying ligand binding sites (pockets) is a foundational problem in structure-based drug design. Classical approaches used geometric cavity detection, evolutionary conservation, and energy-based scoring, and more recent methods apply deep learning with increasing sophistication. However, most methods are still limited by their dependence on input structures. Recent advances in molecular structure prediction provide an opportunity to leverage the rich representations learned by co-folding models to directly predict pockets for a given protein sequence. AF2Bind~\citep{gazizov2023af2bind} showed that the trunk features from AlphaFold2 can be used to predict binding sites, and labs building AF3-era co-folding models also claim impressive results on the blind pocket prediction task~\citep{isodde2026, anewlabs2026anewdde}. However, the field lacks open benchmarks to test these claims, as existing benchmarks predate the training cutoffs of AF3-era co-folding models.

\begin{figure}[t]
\centering
\newcommand{\figonetitle}[1]{{\sffamily\bfseries\fontsize{8}{9.5}\selectfont #1\par}\vspace{4pt}}
\begin{minipage}[t]{0.28\textwidth}\centering
  \figonetitle{A.\ \method{} Method Overview}
  \includegraphics[scale=0.5357]{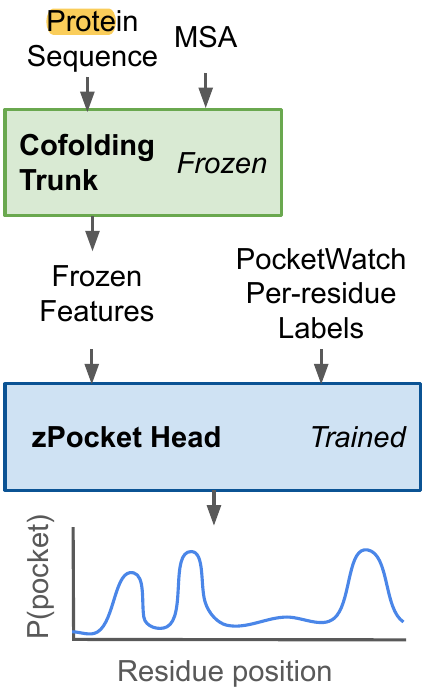}
\end{minipage}\hfill
\begin{minipage}[t]{0.32\textwidth}\centering
  \figonetitle{B.\ Q96SW2 (CRBN) Example}
  \includegraphics[width=102.8pt]{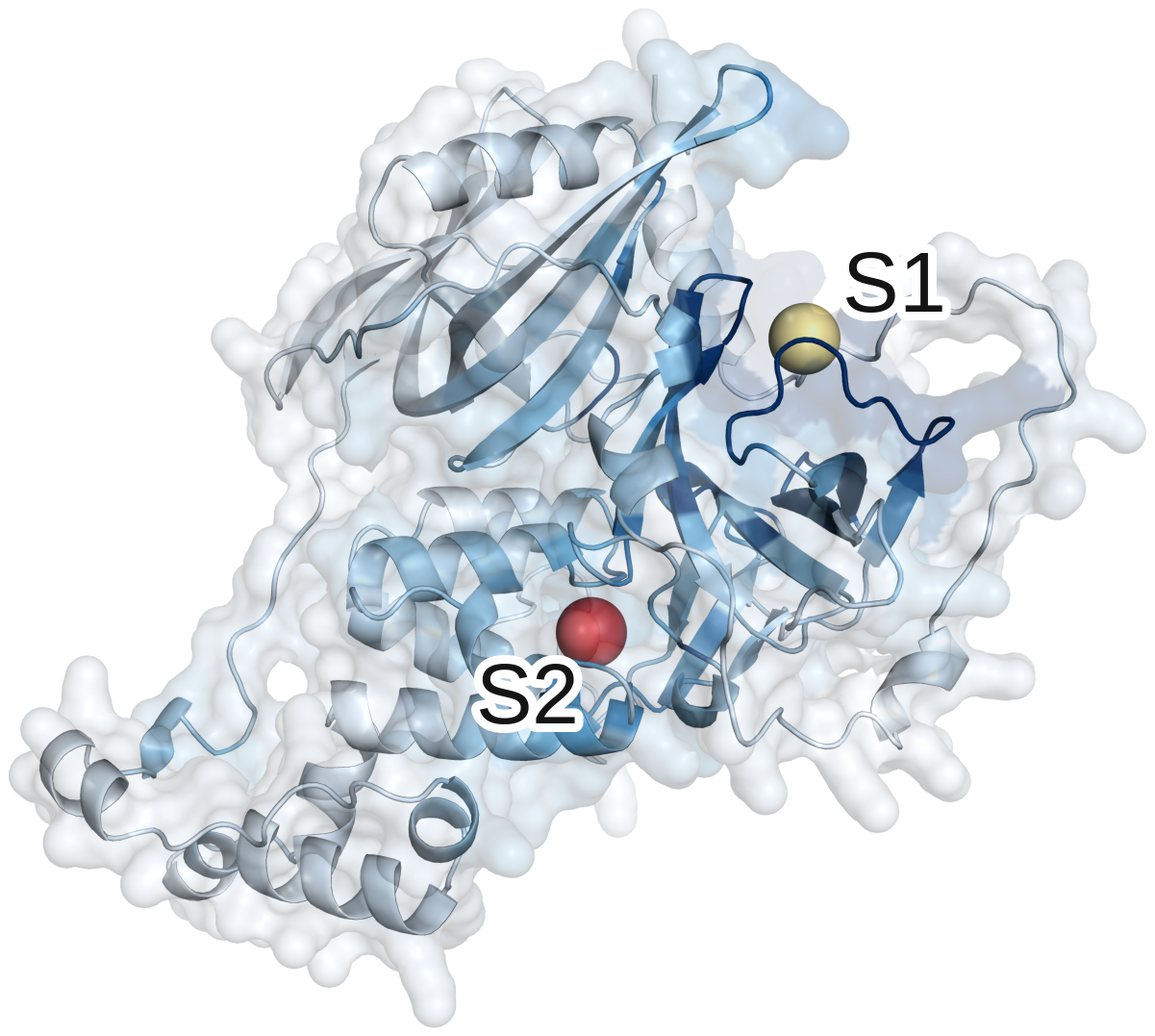}\par\vspace{3pt}
  \includegraphics[width=\linewidth]{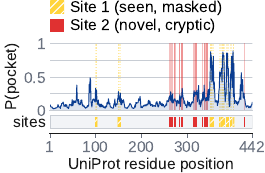}
\end{minipage}\hfill
\begin{minipage}[t]{0.36\textwidth}\centering
  \figonetitle{C.\ \method{} vs.\ Baselines\\on strata of PocketWatch}
  \includegraphics[width=\linewidth]{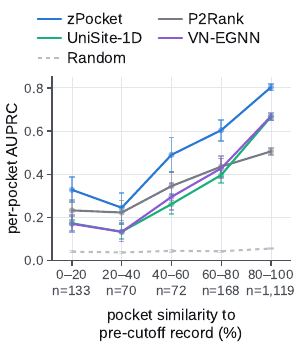}
\end{minipage}
\caption{\textbf{\method{} predicts pocket residues from frozen co-folding features and beats
baselines across different stratifications of the PocketWatch test set.}
\textbf{(A)} Method. A sequence is folded by a frozen trunk and its frozen trunk features are cached. A head is trained on these features and per-residue labels from the PocketWatch dataset, emitting a pocket probability for every residue.
\textbf{(B)} One test target, cereblon (UniProt Q96SW2), with the predicted
per-residue score painted on the fold and plotted along the canonical UniProt
sequence; spheres mark each annotated site's ligand centroid. Site $S_1$ is
already in the pre-cutoff record, so its residues are masked and not scored;
$S_2$ is novel and cryptic, first appearing after the cutoff, and is scored in
full. For more details on this visualization, see Appendix~\ref{app:calibration-figure}.
\textbf{(C)} \method{} (ESMFold2 \texttt{pairformer2}) against P2Rank,
UniSite-1D, and VN-EGNN scored by per-pocket \auprc{} against
\qcov{}, the largest fraction of a held-out pocket's residues covered by a
pre-cutoff pocket. UniSite-1D and VN-EGNN are both retrained on PocketWatch; Table~\ref{tab:main} also gives their published weights. \method{} leads every baseline in every bin. Bin sizes are printed on the axis and bars are bootstrap 95\% intervals.
}
\label{fig:overview}
\end{figure}

This work addresses this benchmark gap and evaluates two leading co-folding models for this task. First, we build \textbf{PocketWatch}, a new temporal-split dataset of ligand binding sites. PocketWatch is a curated, per-UniProt consolidation of biologically relevant pockets whose test set contains unseen pockets with no pre-cutoff record, stratified by sequence similarity and structural similarity to co-folding model training data so we can separate generalization from memorization. We further stratify by crypticity, since cryptic pockets are particularly challenging for structure-based pocket prediction. Second, we develop \textbf{zPocket}, a set of per-residue pocket prediction heads over the trunk features of
Protenix~\citep{zhang2026protenixv2} and ESMFold2~\citep{candido2026worldmodel}. Both models had open weights when we ran our experiments, but the Protenix-v2 weights have since moved to a proprietary license. The main text therefore reports ESMFold2 heads, whose trunk remains open-weight, with every structure-reading baseline run on ESMFold2 folds; the two trunks are compared at matched architectures in Table~\ref{tab:full}, where Protenix-v2 scores slightly higher. We benchmark against P2Rank~\citep{krivak2018p2rank}, a widely used competitive baseline~\citep{utges2024comparative}, alongside two learned detectors evaluated using the published weights as well as retrained on PocketWatch: UniSite~\citep{fan2025unisite}, which reads sequence alone, and VN-EGNN~\citep{sestak2024vnegnn}, which reads a structure graph. We find that \method{} outperforms these baselines in every stratum. Figure~\ref{fig:overview} summarizes the method, provides an illustration of zPocket outputs as well as the per-residue masking the temporal split rests on, and the headline stratified comparison.

\section{Related work}
\label{sec:related}

\textbf{Binding site prediction.} \hspace{1.5mm}
Classical pocket finders score surface geometry~\citep{connolly1983surface,hendlich1997ligsite}, evolutionary conservation~\citep{capra2009concavity}, or interaction energetics~\citep{goodford1985grid,laurie2005qsitefinder}.
Since fpocket~\citep{leguilloux2009fpocket} made cavity enumeration cheap, much of the progress has come from learning to rank cavities rather than to find them: P2Rank~\citep{krivak2018p2rank} trains a random forest to score solvent-accessible surface points from local physicochemical descriptors and clusters the high-scoring points into ranked pockets, while DeepPocket~\citep{aggarwal2022deeppocket} leaves fpocket's proposals untouched and re-ranks them with a 3D CNN over a grid centered on each cavity. Other deep learning methods predict pockets directly from a voxel grid (DeepSite,~\citealp{jimenez2017deepsite}) or a protein graph (VN-EGNN,~\citealp{sestak2024vnegnn}). Despite increased sophistication, reliance on predefined features and explicit geometric inputs still limits these methods. A recent method, Lacuna, detects pockets across a generated conformational ensemble rather than on the input structure alone~\citep{moore2026lacuna}.

\textbf{Pocket-finding benchmarks.} \hspace{1.5mm}
UniSite~\citep{fan2025unisite} and LIGYSIS~\citep{utges2024comparative} curate
pocket datasets along lines similar to ours, but neither provides a temporal split
allowing for evaluation of AF3-era co-folding models. PocketWatch extends the record of
ligand binding sites found in the RCSB through the end of 2025 to allow for the identification of more novel pockets. Following Runs N' Poses~\citep{skrinjar2025runsnposes}, which showed that co-folding accuracy collapses on ligand pockets far from the training set, we stratify our test set by structural similarity to better evaluate generalization. Because a binding site open in a ligand-bound structure can be absent from the unbound form, identifying ligand sites from static structures can be challenging. The term ``cryptic'' is used to describe such pockets, and
CryptoBench~\citep{skrhak2025cryptobench} formalizes this with an apo--holo
pocket RMSD criterion over quality-controlled experimental pairs, and
IsoDDE~\citep{isodde2026} additionally conditions crypticity on what its
co-folding model saw during training. We borrow from both definitions when stratifying our test set.

\textbf{Structure predictors as feature extractors.} \hspace{1.5mm}
AF2Bind~\citep{gazizov2023af2bind} read binding sites out of AlphaFold2 trunk
features, but had to do so indirectly: AF2 takes no ligand input, so AF2Bind
appends twenty ``bait'' amino acids as small-molecule surrogates and classifies
the pair representation between them and the target. Using light heads trained on frozen trunk features is a commonly used recipe for other tasks, such as affinity prediction in Boltz-2~\citep{passaro2025boltz2}.

\section{Methods}
\label{sec:methods}

\subsection{PocketWatch dataset}
\label{sec:dataset_construction}

\begin{figure}[t]
\centering
\includegraphics[width=\textwidth]{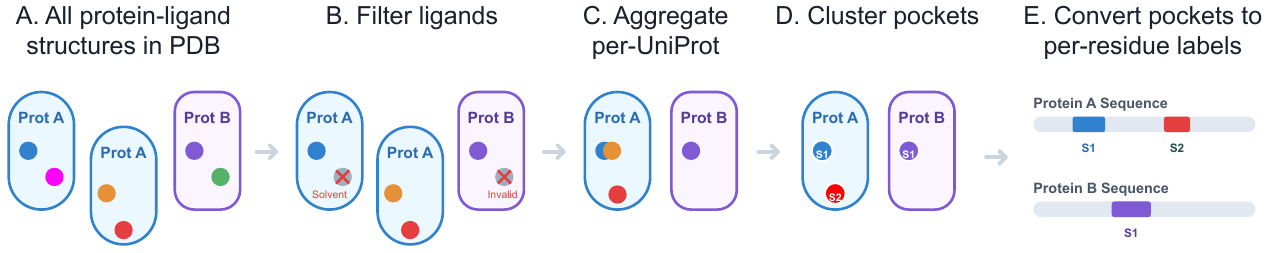}
\caption{\textbf{PocketWatch Construction Overview} \textbf{(A)} We start with every protein--small-molecule structure in the PDB released prior  to January 2, 2026. \textbf{(B)} We filter to only include ligands that are biologically relevant per BioLiP2~\citep{zhang2024biolip2} and pass our size filter; more details in Appendix~\ref{app:dataset}.
\textbf{(C)} Surviving pocket instances (one receptor chain paired with one ligand occurrence) are aggregated per canonical UniProt sequence, so a protein carries every pocket ever observed on it rather than one per structure. \textbf{(D)} Instances are placed in a common frame on a
predicted fold and single-linkage clustered by ligand centroid into discrete pocket sites. \textbf{(E)} Sites become a per-residue binary label vector indexed by canonical UniProt position, which is what the heads are
trained and scored on.}
\label{fig:construction}
\end{figure}

Figure~\ref{fig:construction} provides an overview of PocketWatch dataset construction. We aggregate ligand-contacting residues (``pocket instances'')
from the RCSB on a per-UniProt basis and cluster them into pocket sites.
For each canonical UniProt sequence in our dataset (a ``chain''), we generate a per-residue binary mask based on whether that residue participates in any pocket sites. We then generate a temporal split by assigning each chain to train, validation, or test based on the release date of its earliest member pocket instance. A chain is assigned to test when any of its sites first appeared on or after 2024-01-01, and to validation when any first appeared after 2021-09-30. More details are in Appendix~\ref{app:dataset}.

\textbf{Test set stratification.} \hspace{1.5mm}
We stratify the test set three ways. Test chains are binned by sequence identity to anything the models have already seen, computed with MMseqs2~\citep{steinegger2017mmseqs2}. Test pockets are binned by structural similarity using PLINDER-style maximum pocket coverage
(\qcov{})~\citep{durairaj2024plinder}. Finally, we label crypticity following CryptoBench~\citep{skrhak2025cryptobench} and IsoDDE~\citep{isodde2026}, with additional annotation of nominally cryptic pockets for which the predicted fold and ground truth pocket are actually close in RMSD. Details are in Appendix~\ref{app:strat}.

\subsection{\method{}: trunk-feature-based pocket prediction}
\label{sec:zpocket-method}
\textbf{Training \method{} per-residue heads.} \hspace{1.5mm}
Each protein is featurized once by folding its sequence with two (frozen) co-folding
models: $s$ and $z$ features are retained for Protenix-v2~\citep{zhang2026protenixv2} and
last-layer ESMC-6B embeddings and $z$ are retained for ESMFold2~\citep{candido2026worldmodel}.

We train architectures of increasing expressivity on these frozen features to locate where the pocket signal lives. Every $z$-based head maps a protein's pair representation $z$ to per-residue pocket logits through a common pooling: each residue is represented by the concatenation of its row mean, column mean and self-pair diagonal, which a LayerNorm and a small MLP then reduce to a scalar. The $z$-based heads differ in how much pair mixing precedes that pooling. \textbf{pooled} applies it directly, with no mixing. \textbf{axial} projects $z$ to a narrow width and applies a row-then-column self-attention block. \textbf{pairformer}$K$ applies $K$ AlphaFold-style triangular pair blocks~\citep{jumper2021alphafold}, which let a residue's score depend on geometry it does not directly contact. We also train \textbf{single} heads, which apply that readout MLP directly to a per-residue representation (Protenix $s$ or last-layer ESMC-6B embeddings). Heads are trained with per-residue binary cross-entropy loss and we select the checkpoint with the best per-target validation \auprc{}. Further details are in Appendix~\ref{app:model}.

\textbf{Decoding to pocket centroids.} \hspace{1.5mm}
Ligand binding site prediction is often evaluated using predicted pocket centroids.
We obtain pocket centroids from our per-residue predictions in two ways: \textbf{clustering} and \textbf{re-ranking} centroids from other pocket predictors. The
\textbf{clustering} decoder takes residues scoring above
a threshold to seed a weighted mean shift over their
C$\alpha$ coordinates, and each resulting density mode is reported as one pocket.
The \textbf{re-ranking} decoder instead keeps another method's proposals and
re-orders them, following the division of labor in
DeepPocket~\citep{aggarwal2022deeppocket} but with the \method{} head in place
of its 3D-CNN: each proposed centroid is scored by aggregating the head's probabilities over its lining residues. We also evaluate ``hybrid'' decoders, where we combine the zPocket centroid proposals with the base method's centroid proposals before re-ranking. All decoders, and the hyper-parameters
chosen on validation, are detailed in Appendix~\ref{app:model}.

\subsection{Baselines}
\label{sec:baselines}

We compare against five pocket finders, all run on the same ESMFold2-predicted folds as
\method{}: the geometric detectors \textbf{fpocket}~\citep{leguilloux2009fpocket} and
\textbf{P2Rank}~\citep{krivak2018p2rank} (alphafold configuration), two variants of the ensemble detector
\textbf{Lacuna}~\citep{moore2026lacuna}, which detects pockets across a generated conformational
ensemble rather than on the input fold alone and ranks them with its own fitted model, the
graph-based detector \textbf{VN-EGNN}~\citep{sestak2024vnegnn}, and the set predictor
\textbf{UniSite-1D}~\citep{fan2025unisite}, of which we run the sequence-only 1D model.
We run the baseline methods using predicted folds rather than experimental structures; in Appendix~\ref{app:apo-check} we show that this has no significant effect on P2Rank's per-target performance.

We additionally retrain VN-EGNN and UniSite-1D from scratch with the PocketWatch
train/val split and report the re-trained versions (\textbf{PW}) alongside their published weights (\textbf{pub});
fpocket, P2Rank and Lacuna appear as published only. Since the published weights are fitted to a
different dataset with different ligand policies and contact thresholds, the retrained models allow
us to measure the value of the PocketWatch dataset itself. Retraining required no new annotation;
both models' supervision signals --- per-residue ligandability, and one centroid per pocket for
VN-EGNN's position loss --- are already carried by our labels. The published and re-trained VN-EGNN models differ slightly
in input features, since the authors' current graph builder adds a residue-type one-hot the
published checkpoint lacks. Neither VN-EGNN nor UniSite-1D emits per-residue scores natively --- VN-EGNN emits ranked
points and UniSite-1D a set of scored masks --- so both are reduced to a per-residue field as
described in Appendix~\ref{app:baselines}, which also gives retraining protocols, versions and
settings for every baseline.

\subsection{Evaluation}
\label{sec:metrics}

We evaluate two tasks: per-residue prediction and pocket-centroid prediction. The \textbf{per-residue} task ranks a chain's residues by predicted pocket probability. We score it both per target, treating all of a protein's pocket residues as positives, and per pocket, taking one pocket's residues as the positives and masking out every other pocket. We report average precision (\auprc{}) as the headline, alongside \textbf{AUROC} and \textbf{R-precision}. The \textbf{pocket-centroid} task scores a ranked list of predicted centroids.
Following the top-$N{+}2$ recall proposed by \citet{utges2024comparative}, we cut the list at the protein's own site count plus two and report \textbf{DCC$_{N+2}$}: the fraction of ground-truth sites with a kept centroid within 4\,\AA{} of their ligand centroid. A budgeted recall says nothing about what a method emits beyond the budget, so we also report \textbf{AP} over the whole list and the \textbf{ceiling}, the same DCC rate with no budget. Full definitions are in Appendix~\ref{app:metrics}.

Which pockets are scored is decided once, at the site level, and both tables read the same
decision. We mask ``seen'' pocket sites, those with a member pocket instance predating the test
cutoff; ``interface-only'' pockets, those every one of whose member pockets involves more than one
chain, so that P2Rank and the other single-chain methods are not charged for contacts a monomer
fold cannot express; and the few novel sites whose residues lie entirely inside an already-public
pocket. A pocket observed at an interface in some structures and on a single chain in others is
\emph{not} interface-only: it is scored, on its monomer-expressible residues alone. Masked pockets
are neither credited nor charged --- their residues are excluded from every other pocket's
negatives, and a predicted centroid landing on one is dropped from the ranked list rather than
counted as a false positive. Of \ntestpocketstage{} test pockets surviving the first two rules,
\testpockets{} are scored over \testchains{} chains; the tables report \npockets{} over
\ntargets{}, restricted to the proteins every method returned a prediction for.
\section{Results}
\label{sec:results}

\subsection{\method{} beats all baselines on per-residue pocket prediction}
\label{sec:results-perresidue}

\begin{table}[t]
\centering
\small
\setlength{\tabcolsep}{1.2pt}
\caption{Per-residue results on PocketWatch, for \method{} heads on the \textbf{ESMFold2} trunk against both published and re-trained baselines, scored on the 1{,}562 test pockets of 1{,}275 targets in the effective test set. All baselines are run on ESMFold2's predicted folds when structure is needed. Table~\ref{tab:full} compares results from both ESMFold2 and Protenix. The \texttt{single} row reads the final-layer ESMC-6B embedding. Rp = R-precision. Per-target metrics are computed once per target from that target's residue ranking, then averaged over targets with equal weight. Per-pocket metrics are computed once per pocket, with positives from other pockets masked out, and averaged over pockets with equal weight. The last two columns split AUPRC$_{p}$ by the number of scored pockets for each protein. (pub) is the published weights; (PW) is the same architecture retrained on PocketWatch. UniSite-1D (PW) is the mean of three seeds; every other row is a single model. Bold = best per column.}
\label{tab:main}
\begin{tabular}{llcccccccc}
\toprule
& & \multicolumn{3}{c}{per target} & \multicolumn{3}{c}{per pocket} & \multicolumn{2}{c}{AUPRC$_{p}$ by multiplicity} \\
\cmidrule(lr){3-5}\cmidrule(lr){6-8}\cmidrule(lr){9-10}
Method & Variant & AUPRC$_{t}$ & AUROC$_{t}$ & Rp$_{t}$ & AUPRC$_{p}$ & AUROC$_{p}$ & Rp$_{p}$ & 1 pocket & $>$1 \\
\midrule
Baselines & P2Rank & 0.477 & 0.859 & 0.449 & 0.455 & 0.855 & 0.425 & 0.466 & 0.430 \\
 & VN-EGNN (pub) & 0.559 & 0.862 & 0.513 & 0.516 & 0.846 & 0.472 & 0.567 & 0.409 \\
 & UniSite-1D (pub) & 0.531 & 0.878 & 0.507 & 0.497 & 0.870 & 0.474 & 0.533 & 0.421 \\
\midrule
PocketWatch & VN-EGNN (PW) & 0.598 & 0.837 & 0.567 & 0.559 & 0.819 & 0.530 & 0.605 & 0.463 \\
Retrained & UniSite-1D (PW) & 0.595 & 0.898 & 0.568 & 0.552 & 0.883 & 0.529 & 0.598 & 0.458 \\
\midrule
zPocket & single (ESMC) & 0.611 & 0.918 & 0.564 & 0.574 & 0.906 & 0.528 & 0.610 & 0.499 \\
 & \texttt{pooled} & 0.639 & 0.924 & 0.589 & 0.600 & 0.913 & 0.551 & 0.640 & 0.516 \\
 & \texttt{axial} & 0.716 & 0.945 & 0.652 & 0.675 & 0.934 & 0.613 & 0.719 & 0.584 \\
 & \texttt{pairformer1} & 0.737 & 0.950 & 0.677 & 0.695 & \textbf{0.941} & 0.637 & 0.741 & \textbf{0.600} \\
 & \texttt{pairformer2} & \textbf{0.745} & \textbf{0.951} & \textbf{0.682} & \textbf{0.702} & \textbf{0.941} & \textbf{0.639} & \textbf{0.751} & 0.599 \\
\midrule
\multicolumn{2}{l}{random} & 0.049 & 0.500 & 0.049 & 0.051 & 0.500 & 0.051 & 0.053 & 0.048 \\
\bottomrule
\end{tabular}
\end{table}

Table~\ref{tab:main} reports \method{} head architectures on ESMFold2 trunk features against baselines run on ESMFold2 folds; Table~\ref{tab:full} adds the same heads on the Protenix trunk. Every $z$-based \method{} head on every trunk beats the published and re-trained baselines by a wide margin. The architecture ordering $\texttt{pairformer2} > \texttt{pairformer1} > \texttt{axial} > \texttt{pooled}$ is identical on both trunks and on nearly every metric: pair mixing before the pooled readout buys real signal, but much of the signal can be read directly off $z$ as shown by \texttt{pooled}. The best model, ESMFold2 \texttt{pairformer2}, reaches 0.745 per-target \auprc{} against 0.477--0.598 for the baselines. Table~\ref{tab:full} shows that the two trunks are close to interchangeable at matched architectures, with Protenix slightly but consistently beating ESMFold2. Performance mostly saturates after one pairformer block, the second buying $+0.008$. AUROC spans 0.924--0.951 across the four $z$ heads against 0.837--0.898 for the baselines; at a \testposrate{}\% positive rate it is dominated by easy negatives and does not really separate the heads.

\textbf{The pair tensor is where the signal lives.} \hspace{1.5mm} The \texttt{single} row in Table~\ref{tab:main}
applies the same readout to ESMFold2's final-layer ESMC-6B embedding. Despite receiving no structural input whatsoever, it scores 0.611 and clears every baseline, including VN-EGNN (PW) at 0.598, yet still falls short of the \texttt{pooled} row's 0.639. The ESMC embedding is a much stronger per-residue representation than the $s$ features from Protenix-v2 (Table~\ref{tab:full}): the $s$-feature-based head for Protenix-v2 scores 0.474 against its \texttt{pooled} row's 0.669, so most of its pocket signal sits in $z$ rather than $s$, and it is the one \method{} configuration that does not beat P2Rank on the same folds (0.503).

\textbf{Retraining on PocketWatch improves baselines.} \hspace{1.5mm} VN-EGNN gains
$+0.039$ per-target \auprc{} over its published weights (0.559 to 0.598) and UniSite-1D $+0.064$
(0.531 to 0.595). This suggests that the PocketWatch benchmark is worth training on rather than only scoring against, and it holds for both a graph network reading structure and a transformer reading sequence. Both retrained baselines still fall short of every $z$-based \method{} head, so the trunk representation may still buy something the retraining does not. However, we acknowledge that some differences in training procedures (Appendix~\ref{app:baselines}) may also affect results, so it's not possible to entirely attribute the gains to the new data without further ablations. We also note that the PocketWatch training set is more closely matched to the PocketWatch test set than the training distributions of published methods, which is another mechanism by which retraining can improve performance.

\textbf{Multi-pocket proteins are harder.} \hspace{1.5mm} The last two columns in Table~\ref{tab:main} split
per-pocket \auprc{} by how many scored pockets a chain holds. Every $z$ head
does worse on chains holding more than one, with \texttt{pairformer2} falling from 0.751 to 0.599.

\begin{figure}[t]
\centering
\includegraphics[width=0.95\textwidth]{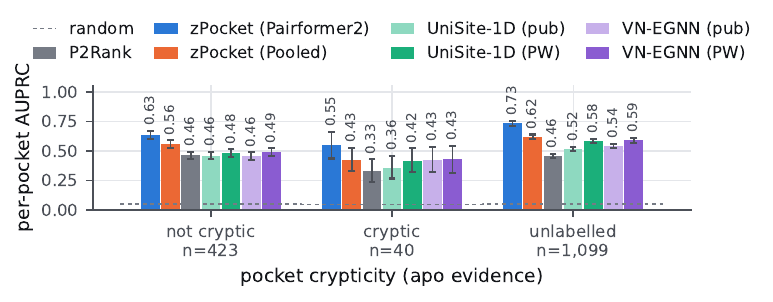}
\caption{Per-pocket \auprc{} by crypticity label. \emph{Cryptic} pockets are
those that an experimental apo structure shows closed; \emph{not cryptic} are those an experimental apo structure shows open;
\emph{unlabeled} those with no admissible experimental apo structure, which is 70\% of the test set and not a neutral remainder; a crypticity label requires a deposited apo structure, which makes the
labeled subset a harder, biased sample. Bars are bootstrap 95\% intervals; the cryptic column holds 40 pockets and its intervals are wide accordingly. Appendix~\ref{app:crypticity-results}
crosses this with predicted-fold accuracy.}
\label{fig:cryptic-coarse}
\end{figure}

\textbf{Pocket prediction drops for unfamiliar pockets.} \hspace{1.5mm} To test whether zPocket is re-identifying pockets from co-folding training data, we stratify the test set by \qcov{}, the largest fraction of a held-out pocket's residues covered by a pre-cutoff pocket. Per-pocket \auprc{} falls from 0.803 in the most-covered bin to 0.327 below 20\% coverage (Fig.~\ref{fig:overview}C), and 1{,}119 of \npockets{} scored pockets --- \toppocketpct{}\% --- sit in that top bin, so the aggregate in Table~\ref{tab:main} is dominated by familiar pockets. Structural similarity is the more informative leakage axis compared to sequence similarity: per-target \auprc{} is roughly flat across sequence identity until the $\geq$95\% bin, where it falls (Fig.~\ref{fig:sequence}). We read that fall as an artifact of seen pockets being masked, leaving only harder pockets. Despite lower results on unfamiliar pockets, \texttt{zPocket pairformer2} still leads baselines in every similarity stratum. In Figure~\ref{fig:cryptic-coarse}, we compare results across the crypticity stratification, once again observing that \method{} beats the baseline methods in every stratum, but we note the small number of cryptic pockets. Appendix~\ref{app:crypticity-results} crosses crypticity with predicted-fold accuracy and shows that most of the apparent effect is the fold's, not the pocket's.
\subsection{The pocket signal is not specific to one trunk}
\label{sec:results-trunks}

\begin{table}[t]
\centering
\small
\caption{Per-target \auprc{} of \method{} heads trained with one recipe on seven frozen co-folding
trunks, scored on the \npockets{} test pockets of \ntargets{} targets as in Table~\ref{tab:main}.
$c_z$ is the trunk's pair width. Brackets are 95\% bootstrap intervals over targets. The
Protenix-v2 and ESMFold2 rows repeat Table~\ref{tab:full}. Each row is a single training run.
Below the double rule, ensembles average the per-residue logits of those \texttt{pairformer2}
heads with equal weight; nothing is fitted. Bold = best single trunk per column. Details in
Appendix~\ref{app:trunks}.}
\label{tab:trunks}
\begin{tabular}{lccc}
\toprule
Trunk & $c_z$ & \texttt{pooled} & \texttt{pairformer2} \\
\midrule
Protenix-v2       & 256 & \textbf{0.669} [0.653, 0.684] & \textbf{0.757} [0.741, 0.772] \\
Chai-1            & 256 & 0.668 [0.652, 0.683] & 0.746 [0.729, 0.762] \\
ESMFold2          & 256 & 0.639 [0.623, 0.655] & 0.745 [0.729, 0.761] \\
Protenix-v1       & 128 & 0.650 [0.634, 0.665] & 0.742 [0.725, 0.758] \\
OpenFold3-preview & 128 & 0.619 [0.604, 0.634] & 0.742 [0.726, 0.758] \\
RF3               & 128 & 0.631 [0.615, 0.646] & 0.741 [0.725, 0.757] \\
Boltz-1           & 128 & 0.638 [0.622, 0.653] & 0.733 [0.716, 0.749] \\
\midrule\midrule
Protenix-v2 + ESMFold2 & --- & --- & 0.773 [0.757, 0.788] \\
All seven trunks       & --- & --- & 0.777 [0.762, 0.793] \\
\bottomrule
\end{tabular}
\end{table}

To test whether the pocket signal in $z$ is particular to ESMFold2 and Protenix-v2, we trained z-pocket heads on features from 5 more co-folding trunks with openly released weights: Protenix-v1~\citep{protenix2025},
OpenFold3-preview~\citep{openfold2025of3}, Boltz-1~\citep{wohlwend2024boltz1},
Chai-1~\citep{chai2024chai1} and RF3~\citep{corley2025rf3} using the procedure described in 
Section~\ref{sec:zpocket-method}. Every checkpoint's experimental training structures
predate the PocketWatch validation and test periods. Featurization and training details are in
Appendix~\ref{app:trunks}.

All of the co-folding trunks carry the pocket signal (Table~\ref{tab:trunks}): each \texttt{pairformer2} head scores
between 0.733 and 0.757, well clear of the strongest baseline in Table~\ref{tab:main} (0.598). The head architecture matters more than the trunk: \texttt{pairformer2} beats \texttt{pooled} by 0.08--0.12 on every
trunk. The seven trunks span 0.024 under \texttt{pairformer2}, half their 0.050 spread under
\texttt{pooled}, suggesting that pair mixing recovers signal some trunks do not expose to a pooled readout. OpenFold3-preview is the clearest case, last with \texttt{pooled} and level with ESMFold2
with \texttt{pairformer2}. Protenix-v2 scores highest with both heads, $+0.015$ over its
predecessor Protenix-v1 with \texttt{pairformer2} (paired 95\% CI $[+0.007, +0.023]$), though
Chai-1 ties it with \texttt{pooled} ($-0.001$ $[-0.006, +0.004]$).
Ensembling trunks helps with no tuning: averaging the \texttt{pairformer2} heads' logits,
Protenix-v2 with ESMFold2 reaches 0.773 ($+0.016$ $[+0.011, +0.021]$ over Protenix-v2 alone) and
all seven trunks 0.777 ($+0.021$ $[+0.015, +0.027]$). As each head is a single seed, part of this gain may come from averaging independent training runs rather than from the trunks' diversity.

\subsection{Pocket centroids}
\label{sec:results-discrete}


\begin{figure}[t]
\centering
\includegraphics[width=\textwidth]{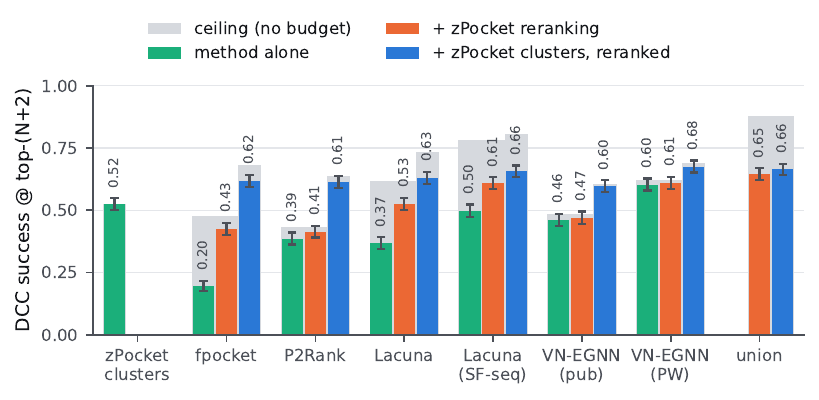}
\caption{\textbf{Pocket centroid prediction on PocketWatch.} DCC$_{N+2}$ over \npockets{} pockets
of \ntargets{} targets, at 4\,\AA{}. Green is each method's own proposals under its own ranking;
orange re-ranks those same proposals with the \method{} head; blue adds the \method{} cluster
proposals to them before re-ranking with the \method{} head. Behind \emph{each} bar in gray is that bar's own ceiling --- the total number of true pockets detected among all of the proposals. A bar close to its ceiling is limited by what was proposed; a short bar under a tall ceiling is limited by the ranking. \emph{union} pools every method's cavities (other than zPocket clusters), so it has no ranking of its own. Error bars are 95\% percentile bootstrap intervals
over pockets and are \emph{unpaired}: two bars whose intervals overlap can still differ
significantly once the comparison is paired. By a \emph{paired} 95\% bootstrap over the same
pockets, re-ranking lifts every method except for the PocketWatch-retrained variant of VN-EGNN (PW),
and combining the \method{} clusters with any other method's proposals then re-ranking lifts
DCC$_{N+2}$ further.}
\label{fig:treatments}
\end{figure}

In Figure~\ref{fig:treatments}, we compare different pocket proposal and pocket ranker combinations in terms of DCC$_{N+2}$, the fraction of true pocket centroids detected among the top ranked $N+2$ predicted centroids, where $N$ is the number of true pockets for a given protein; we also show the ceiling, which is the total percentage of true pocket centroids detected among all pocket proposals, and therefore the upper bound on DCC$_{N+2}$. Table~\ref{tab:discrete} gives the same results numerically along with average precision, precision, and emission rate for each method.

\textbf{zPocket's clustering is precise but limited.} The \method{} clustering decoder emits only 1.1 pockets per protein on average, so cannot reach a protein's second site and its ceiling is 0.524. It is the most precise method by a wide margin --- 60\% of what it emits lands on a real site, against 18\% for retrained VN-EGNN and 4\% for P2Rank --- but its recall is limited by its low emission rate. Despite this limitation, the \method{} clustering decoder is competitive with the other methods in terms of DCC$_{N+2}$, beating all published baselines using their published rankers, and only falling short of the PocketWatch-retrained VN-EGNN method.

\textbf{Re-ranking with \method{} improves ranked pocket recall.} Re-ranking a method's pocket proposals with \method{} significantly improved each DCC$_{N+2}$ for each published baseline (comparing orange bars to green bars), and only failed to improve the PocketWatch-retrained VN-EGNN method. Re-ranking increased DCC$_{N+2}$ by an average of 0.090 across these 6 methods, with the biggest gains for fpocket and the two Lacuna variants.

\textbf{\method{} contributes proposals, not just ranking.}
We see further improvements when we combine the \method{} clustering proposals with each method's own proposals and then  re-rank the union with the \method{} head (blue bars). In fact, every method, even VN-EGNN (PW), sees significant improvement in DCC$_{N+2}$ when combined with zPocket proposals, though we note this is not the case for DCC$_{N}$ (Table~\ref{tab:discrete}). Across these six methods and the union method, the average DCC$_{N+2}$ gain from adding \method{} proposals and re-ranking is 0.108. We note that the highest-scoring method is the re-trained VN-EGNN method combined with \method{} proposals, which reaches 0.675. We note that even though the union of proposals had a higher ceiling, we were not able to improve DCC$_{N+2}$ beyond this through ranking.

\textbf{Pocket detection deteriorates on less familiar pockets, and ranking gets harder.} \hspace{1.5mm} In Appendix~\ref{app:centroid-results}, we split this comparison by \qcov{}, and observe that the ceiling and DCC$_{N+2}$ both drop for less-familiar pockets; In the $\geq$80\% bin the best configurations reach DCC$_{N+2}$ of about 0.81, while below
80\% coverage no configuration exceeds 0.37. Furthermore, the gap between the ceiling and the ranked DCC$_{N+2}$ is larger for less-familiar pockets, indicating that ranking is also more difficult.

\section{Applications}


\subsection{In-Silico Fragment Screening}

\textbf{Evaluation dataset.} \hspace{1.5mm} Fragment screening campaigns are used to find preliminary
hits against a protein target and characterize possible pockets. The protein is screened against a
library of several hundred fragments and a structure is solved for every hit. This process is time
consuming and expensive, and ML pocket predictors hold promise in reducing costs through narrower
screening of fragments or progressing directly to hit ID. To simulate a fragment screening scenario,
we assembled a test set from eleven XChem campaigns released through Fragalysis together with four
fragment-screen group depositions in the PDB (details in Appendix~\ref{app:fragment-screening}). A
fragment is counted once per distinct SMILES, and sites are single-linkage clusters as in
PocketWatch. We score a site only if at least three distinct fragments support it, its hits are not
dominated by crystal contacts (median symmetry-contact fraction below $0.25$), and its lining
residues map onto the predicted fold with a C$\alpha$ fit below $2$\,\AA{}. Since that last check is
run against a predicted structure, we require it to pass on \emph{both} Protenix-v2's and ESMFold2's
folds, so that the two are scored on identical labels. Sites whose lining residues overlap with a
pocket in the \method{} heads' training split are dropped. This leaves $12$ proteins with $35$ scored
sites.

\begin{figure}[t]
\centering
\includegraphics[width=\textwidth]{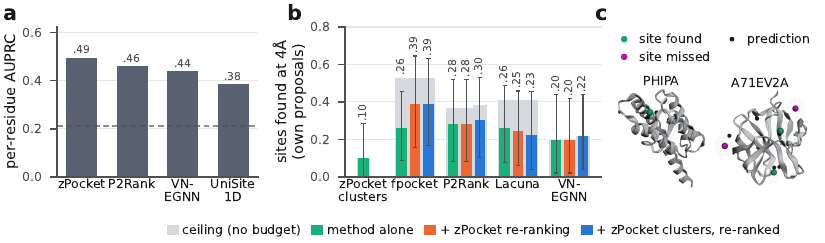}
\caption{\textbf{Pocket prediction on crystallographic fragment screens.}
\textbf{(a)} Per-residue AUPRC over the $35$ fragment-soak sites of $12$ proteins, for the \method{}
head on ESMFold2 folds against three baselines, also using ESMFold2 folds when needed; the dashed line is
chance.
\textbf{(b)} The same proteins scored as in Figure~\ref{fig:treatments}: each detector's own
proposals under its own ranking, those proposals re-ranked by the \method{} head, and those proposals
pooled with the head's cluster proposals before re-ranking. A site counts as found when a prediction
in the top $N+2$ lies within $4$\,\AA{} of its centroid, $N$ being that protein's scored-site count.
Behind each bar in gray is that bar's own ceiling, the rate with no budget at all. Error bars are $95\%$ percentile bootstrap intervals over proteins, and they are wide: twelve proteins separate none of these methods.
\textbf{(c)} Two targets scored by fpocket's cavities pooled with \method{} clusters and re-ranked.
Spheres mark each scored site's centroid, drawn at the $4$\,\AA{} success radius; the small
black spheres are the predicted centroids inside the budget. PHIPA carries one scored site and it is
found; A71EV2A carries four and two are found. Figure~\ref{fig:fragsoak-protenix} repeats (a) and
(b) using the Protenix-v2-based head and folds.}
\label{fig:fragsoak}
\end{figure}

\textbf{The per-residue ordering holds, but re-ranking helps unevenly.} \hspace{1.5mm} On per-residue
scores the PocketWatch ordering holds: \method{} reaches $0.494$ \auprc{} against $0.459$ for P2Rank,
$0.438$ for VN-EGNN and $0.385$ for UniSite-1D, all far above the $0.212$ a random score reaches
(Figure~\ref{fig:fragsoak}a). Turning those scores into centroids is harder, and the gain from
re-ranking varies by detector. Re-ranking fpocket's cavities with the \method{} head lifts it from
$0.259$ to $0.391$, the best configuration measured (Figure~\ref{fig:fragsoak}b); pooling the head's
own cluster proposals in before re-ranking adds nothing further, since on this trunk the clusters
reach no site fpocket's cavities do not. Re-ranking leaves P2Rank and VN-EGNN mostly unchanged,
though adding the \method{} cluster proposals gives both a slight boost. Lacuna drifts down under
both treatments, $0.263$ to $0.246$ re-ranked and to $0.225$ pooled. On the Protenix-v2 trunk the \method{} cluster centroids do add sites the
detectors miss, and pooling them in lifts both fpocket and Lacuna (Appendix~\ref{app:fragsoak-protenix}). With twelve proteins the intervals are wide (fpocket
re-ranked spans $[0.16, 0.65]$), so no pair of configurations here separates; we read this set as
corroborating the PocketWatch comparison rather than ranking methods on its own.

\textbf{Qualitative differences from PocketWatch.} 
\hspace{1.5mm} Two effects may explain why
DCC$_{N+2}$ on the fragment screening set is lower than on the PocketWatch test set. At the
$4$\,\AA{} radius PocketWatch is scored at, the best configuration here reaches $0.391$
DCC$_{N+2}$, while on the PocketWatch test set it reaches $0.68$. The first is site multiplicity: a
fragment-screened protein carries $2.9$ scored sites on average against PocketWatch's $1.23$, and a
decoder emitting about one pocket per protein cannot reach a third or a fifth site. However,
restricting each protein to its dominant site (the site with the most fragments) lifts $0.391$ only
to $0.417$. Perfect ranking also does not close the gap, since the best proposal set reaches a
ceiling of $0.529$ with no budget at all (Figure~\ref{fig:fragsoak}b), well short of the $0.879$
ceiling on the PocketWatch test set. The second, and perhaps more important factor, is that
fragment-screening pockets are different from drug-like pockets, as fragments are often smaller and
can have much lower affinity than true binders. Therefore, it is possible that fragment soaking and
\method{} offer complementary information for identifying pockets.

\subsection{Novel pocket hypotheses}

\textbf{Searching for novel pockets.} \hspace{1.5mm} Strong pocket predictors can also be used to generate new ideas in target selection by surfacing high-confidence, but previously unexplored, pockets on proteins with therapeutic potential. As a proof-of-concept, zPocket was run across the $\sim$20,000 reviewed human proteins of UniProt proteome
UP000005640, yielding $\sim$22,000 predicted pockets. Unlike the rest of the main text, this scan uses the Protenix-v2 \texttt{pairformer2} head on Protenix-v2 folds; it was run while the Protenix-v2 weights were openly licensed and has not been repeated on ESMFold2. The resulting pockets were then filtered for basic geometric plausibility (size and buriedness) and novelty. Novelty was determined using both sequence-level UniProt annotations and Q-BioLiP structural annotations. These annotations were used to check the input protein directly (first degree) and analogous proteins (second degree), as judged by shared structural motifs (Folddisco) or shared phylogeny. A pocket that survives these filters is one that would be very unlikely to be found by traditional tools or a database lookup. We describe the full set of novelty filters in Appendix~\ref{app:pocket-filters}.

\textbf{Candidate pocket examples.} \hspace{1.5mm} Of $\sim$22,000 predicted
pockets, 359 pockets on 358 proteins survived novelty and plausibility filters if polymer interface sites are not removed, which accommodates, for example, peptide binding pockets. Discarding these leads to 155 on 154 proteins. Two survivors illustrate possible ideas resulting from large-scale screening and filtering (Appendix~\ref{app:high-conf-pockets}): DET1, a substrate-adaptor subunit of a cullin-RING ligase whose pocket is a candidate
degrader-recruiter site supported by twelve deposited structures, and COP1, which the filter flags as a known peptide-binding interface, and which no geometric detector proposes at all, yet whose predicted pocket covers fourteen of the fifteen residues contacting the bound degron peptide. In this case, zPocket re-identified a peptide-binding site, despite not being trained with peptide ligands. This pocket is not a completely novel pocket but modality-novel, as there is no known small ligand binder despite zPocket's high-confidence prediction. In discovery, there is an obvious tension between novelty and tractability. Predicted pockets on well-characterized proteins are rarely novel, especially on the reference human proteome. This was a first pass with conservative filters, and future work could refine the filtering criteria used, or focus on particular targets. Future work could also extend to screening variants and proteomes for animal models or microbes.

\section{Discussion}
\label{sec:conclusion}

Together these results show that modern co-folding models offer a promising foundation for
blind binding site prediction. \method{} heads outperform every baseline we ran on
per-residue pocket prediction, and improve the rankings of pockets proposed by other
methods. PocketWatch fills a benchmark gap, allowing direct comparison of pocket finders
on pockets that postdate what co-folding models have seen; its training split is also
useful in its own right, lifting both baselines we retrained on it. We hope PocketWatch will be a valuable resource for developers and evaluators of pocket finding methods.

However, challenges remain.
As shown by our stratifications, the number of truly novel pockets in the test set is low, and the aggregate performance is dominated by pockets with a close structural precedent; we expect the number of novel pockets available for evaluation to increase
as more structures are deposited and PocketWatch is extended. We evaluate
monomer pockets only, and extending to interfaces is natural future work, particularly since interface-adjacent sites are where several of our candidate pockets sit.

Finally, we also note that predicted sites may not be functional sites. However, we think this matters less than it once did and may actually be an asset: as degraders, molecular glues, and radioligand conjugates make non-orthosteric sites therapeutically actionable, the useful question shifts from where a protein does its chemistry to where a small molecule can bind at all.


\bibliographystyle{iclr2027_conference}
\bibliography{references}

\appendix

\setcounter{figure}{0}
\setcounter{table}{0}
\renewcommand{\thefigure}{A\arabic{figure}}
\renewcommand{\thetable}{A\arabic{table}}
\renewcommand{\theHfigure}{A\arabic{figure}}
\renewcommand{\theHtable}{A\arabic{table}}

\section{Detailed methods}
\label{app:methods}
This section mirrors Section~\ref{sec:methods}: one subsection per methods
subsection, in the same order.

\subsection{PocketWatch dataset}
\label{app:dataset}
\paragraph{Initial curation}
Following LIGYSIS~\citep{utges2024comparative}, we limit our set of ligand
interactions to those defined ``biologically relevant'' per
BioLiP2~\citep{zhang2024biolip2}. We started from the
Q-BioLiP~\citep{wei2024qbiolip} dataset, as it provided the most up-to-date
snapshot at the time of writing (annotations through 2026-01-02) and also
provides annotations for the number of neighboring chains, making it easy to
restrict pockets to those formed by single chains. We match ligand pocket rows
from Q-BioLiP to their corresponding RCSB ligand pockets, which lets us map each contacting residue onto its underlying UniProt sequence through SIFTS and
annotate every residue with a heavy atom within 5\,\AA{} of the ligand.

Following IsoDDE~\citep{isodde2026}, we make further curation decisions to
restrict our set of pockets to high-quality ``drug-like'' pockets. We define a
\emph{pocket instance} as one receptor chain paired with one ligand occurrence. Its residues are
those with a heavy atom within 5\,\AA{} of a ligand heavy atom, computed from the deposited
coordinates; we also support taking Q-BioLiP's own published contact set instead, which is the stricter of the two, but every number in this paper uses the 5\,\AA{} definition. We retain instances whose ligand has
10--50 resolved heavy atoms (slightly stricter than IsoDDE's 6--50 range), whose chain has $\geq$10 contact residues with
$>$80\% SIFTS~\citep{velankar2013sifts} coverage, and which map to exactly one
UniProt accession of canonical length 40--1200. We then further remove
crystallization additives, metal clusters and metallo-cofactors, tetrapyrroles,
lipids and terpenoids, small metabolites, and free amino acids, while retaining
organic cofactors, saccharides, and nucleotides. We recognize that
curating such a dataset involves a long series of decisions on inclusion/exclusion
criteria, contact-residue definition, and pocket clustering protocol; we have
parameterized each of these decisions to enable different PocketWatch constructions
in the future; here we present results on a first iteration of PocketWatch.

\paragraph{Structural pocket-site clustering.}
A single binding site can be observed many times with different ligands, so we
cluster the resulting set of pocket instances into discrete pocket sites.
Following IsoDDE, we place each pocket in a common frame by fitting
contact-residue C$\alpha$ atoms onto the corresponding positions of a single
Protenix-predicted structure of the canonical UniProt sequence, then carrying
that pocket's observed ligand centroid through the same rigid transform.
We then join pocket instances whose centroids lie within 5\,\AA{} through single-linkage clustering.
Each pocket site is left with a labeled centroid in the frame of the Protenix fold, and
a set of contact residues which is the union of its members' contact residues.
Sites were clustered in this Protenix-fold frame and the clustering was not repeated for ESMFold2,
so site membership is identical for both trunks. For evaluation on ESMFold2, each site's centroid
was re-projected onto the ESMFold2 fold of the same sequence by the same procedure: every member
instance's contact-residue C$\alpha$ atoms are fitted onto the ESMFold2 fold and its ligand
centroid carried through that transform, and the site centroid is the mean over members. 

\paragraph{Per-residue labels.}
We emit a per-residue label vector per protein, indexed by canonical
UniProt position. A residue is positive if it is part of one of its protein's
ligand binding sites, and negative otherwise. Additionally, we provide a
per-residue mask; masked residues are neither positive nor negative at training
and evaluation time, and are dropped from the loss and from every metric.

Some ligand pockets involve multiple contacting chains. Where every member pocket
instance of a site does, we call that site ``interface-only''. Whether these
count as pockets is a policy choice, but we choose to mask them at evaluation
time to enable fair comparison to P2Rank and other baselines, which cannot see the neighboring chains
when run on single-chain folds. Table~\ref{tab:dataset} summarizes the resulting
splits, with and without interface-only pockets.

\begin{table}[t]
\centering
\small
\caption{PocketWatch statistics. Splits are temporal and assigned per protein. The parenthesized numbers show the number of surviving chains/pockets/residues when both the seen and multi-chain-only pockets are masked.
\textbf{Effective test set} is the set every number in this paper is computed on; it applies two further exclusions. First we mask \testsubsumed{} pockets that are novel by centroid but whose residues lie entirely
inside an already-public pocket, leaving \testchains{} chains / \testpockets{} pockets. Second we drop the \ndroppedpockets{} pockets on four proteins that VN-EGNN did not support due to unknown residues.}
\label{tab:dataset}
\begin{tabular}{lrrrr}
\toprule
Split & Chains & Pockets & Residues & Positive Residue \% \\
\midrule
\tabdatasetbody
\bottomrule
\end{tabular}
\end{table}

\subsubsection{Test set stratification}
\label{app:strat}

\paragraph{Sequence-similarity stratification.}
We use MMseqs2~\citep{steinegger2017mmseqs2} to stratify test chains
based on similarity to the union of the trunk's
pretraining sequences and the PocketWatch training and validation sequences. Validation sequences are included in the reference set because a split used for
model selection is not unseen. A hit counts when it spans at least 50 aligned
residues and covers half of \emph{either} sequence, because a pretraining record
may be just a single domain; identity is taken as
the larger of the alignment-length figure and one normalized by the shorter
sequence, since gap columns otherwise penalize a structure with missing unresolved
loops. Independently, we flag any protein whose deposited structure appears in
the pretraining index outright. Both measures err toward reporting more leakage.
Only 16\% of scored test proteins fall below 30\% identity to anything the model has
seen, while 45\% sit at or above 70\% and 32\% at or above 95\%, so a single averaged number
over this test set would substantially overstate generalization.

\paragraph{Pocket-similarity stratification.}
\label{app:qcov}A temporal cutoff guarantees that a held-out site is novel \emph{by date}, not \emph{by structure}: a site whose observations are all recent may sit on a fold
whose homologues were observed with a bound ligand before the cutoff.
Writing $P$ for the residues lining a held-out site's ligand and $Q$ for the
residues lining a ligand in a pre-cutoff structure $T$,
\begin{equation}
\mathrm{qcov}(P, T) \;=\; \frac{\bigl|\,\{p \in P : p \text{ aligns onto some } q \in Q\}\,\bigr|}{|P|},
\label{eq:qcov}
\end{equation}
where the alignment is a structural alignment of the query chains onto $T$'s
chains, and a site's value is the maximum over all pre-cutoff structures as well
as over its own member pocket instances. Equation~\eqref{eq:qcov} is the
pocket-coverage term of the PLINDER~\citep{durairaj2024plinder} similarity
suite. It's the protein half of the Runs N'
Poses~\citep{skrinjar2025runsnposes} train--test metric; we deliberately drop
the Runs N' Poses ligand-shape term, since we ask whether a site has been seen with a
ligand at all rather than whether a specific ligand was seen.

Our setting differs from PLINDER's in one structural respect and three
parameters. PLINDER scores a fixed pair of systems whose chains are already in
correspondence, whereas we search a database and must therefore choose one. We
assign correspondence using a maximum-weight bipartite matching, so that each
pocket chain is mapped to at most one target chain and vice versa, so a
homomeric target cannot accumulate coverage by matching one query chain against
several copies of itself. We take $P$ (the query) as a $5$~\AA\ heavy-atom
shell, matching the contact definition the PocketWatch labels are built at, and
keep PLINDER's $6$~\AA\ for $Q$ (the target): the query radius sets the
denominator and so belongs to the label definition, whereas the target radius
decides what counts as a binding site, so this choice is conservative in that a
wider target shell can only surface more precedent, never less. We consider only
$Q$ residues contacting a non-water, non-polymer ligand of at least six heavy
atoms that is not, following PLINDER, on their ion-and-artifact exclusion list.
Finally, we impose no
sequence-identity floor on hits, unlike the $0.2$ used by Runs N'
Poses~\citep{skrinjar2025runsnposes}, since a remote homologue holding a ligand
in the same site is exactly the precedent we mean to count.

The reference set is a Foldseek~\citep{vankempen2024foldseek} database over all
RCSB~\citep{berman2000pdb} biological-assembly mmCIFs, restricted at build time
to entries whose initial release date falls on or before 2021-09-30
($182{,}401$ of $246{,}905$ entries), so every hit is admissible by construction.
Foldseek retains at most $300$ hits per query chain, and scoring reads at most
$250$ candidate targets per query. We re-scored every read-limited member with
that cap effectively removed and re-searched at
$5{,}000$ hits per chain every member of a site below the $0.8$ threshold whose
query chain was truncated; a site already above the threshold can only rise
within its own bin. Afterwards $48$ sites still retain a chain truncated even at
$5{,}000$, so at most $48$ further sites could belong at $\mathrm{qcov} \ge 0.8$.
We acknowledge that despite conservative choices \qcov{} remains a lower bound
on the precedent that exists; the novel tail of our test
set should therefore be read as an upper bound on how novel it is, and the
metric could be tightened in future iterations of PocketWatch.

\paragraph{Crypticity.}
A site is cryptic when no ligand-free experimental structure presents the pocket
in a conformation resembling its ligand-bound form. For each site we exhaustively
enumerate PDB chains mapping to its UniProt accession, keep those where the site
itself is unoccupied, map residues through SIFTS, superpose each apo chain on
each holo chain over pocket C$\alpha$ atoms, and take the minimum all-atom and
backbone pocket RMSD. Following CryptoBench~\citep{skrhak2025cryptobench} we
require a pair to observe the same pocket residues and both structures to be
$\leq$2.5\,\AA{}, with a 2\,\AA{} threshold; following IsoDDE~\citep{isodde2026}
we combine a resolution-independent criterion with one restricting apo evidence
to structures predating the predictor's training cutoff.

Because any filter that drops apo--holo pairs can only raise a minimum, filtering
can only push a site \emph{toward} cryptic. We therefore make missing evidence
explicit: each criterion is cryptic, refuted, or \emph{unknown}, and sites without
sufficient evidence are reported as unknown rather than defaulted into either
class. Of the \npockets{} scored test sites, 463 carry a label --- 40 cryptic, 423 not cryptic.

The cryptic label describes experimental structures and says nothing about the predicted structure our models consume. Interestingly, we noticed that the predicted apo fold often reproduces the \emph{holo} pocket to within 2\,\AA{} all-atom even for supposedly ``cryptic'' pockets; furthermore, some sites labeled ``not cryptic'' had predicted folds with greater than 2\,\AA{} all-atom RMSD. This is why we cross the cryptic label with
whether the predicted pocket lies within 2\,\AA{} all-atom of the experimental
holo pocket (Fig.~\ref{fig:cryptic}, which carries the per-stratum counts). We observe fold accuracy better explains model performance than the cryptic label itself (Appendix~\ref{app:crypticity-results}). Only
\textbf{17} scored sites are both cryptic and not already open in the ESMFold2
predicted fold, too few for a stable \auprc{}, so we treat them as a case-study. We note that the previously reported results on cryptic pockets by \citet{isodde2026} and \citet{anewlabs2026anewdde} do not report the size of the cryptic subset, and followed similar procedures to ours, so it is possible that their reported results are also based on a small number of cryptic pockets.

\subsection{\method{}: trunk-feature-based pocket prediction}
\label{app:model}

\paragraph{Featurization.}
Each protein is featurized once by folding its sequence with two structure
predictors, Protenix-v2~\citep{zhang2026protenixv2} and
ESMFold2~\citep{candido2026worldmodel},
each with MSAs, and persisting the frozen trunk's internal representations rather
than only the predicted coordinates. Protenix-v2 is run with 10 recycling cycles. ESMFold2 is
run with \texttt{num\_loops}$=10$, which passes the recurrent trunk 11 times before the pair readout.
Per sequence we store, in half precision,
the pairwise representation $z \in \mathbb{R}^{L\times L\times 256}$ and the raw
distogram logits $\in \mathbb{R}^{L\times L\times 64}$, together with a
per-residue single representation $s$ (the pairformer single track,
$\mathbb{R}^{L\times 384}$, for Protenix; the input single embedding,
$\mathbb{R}^{L\times 451}$, for ESMFold2) and, for ESMFold2, the final-layer
ESMC-6B language-model embedding $\in \mathbb{R}^{L\times 2560}$. These
precomputed features are keyed by sequence hash and consumed by the
pocket-prediction heads with the trunks held frozen, so the structure models are
run only once and never fine-tuned.

\paragraph{Head architectures.}
Table~\ref{tab:heads} contains the details of parameter counts for the heads described in Section~\ref{sec:zpocket-method}.

\begin{table}[t]
\centering
\caption{\method{} head architectures. All heads share a common readout and
differ only in the mixing applied to $z$ beforehand.
Parameter counts are for pair width $C=256$ and exclude the frozen trunk.}
\label{tab:heads}
\begin{tabular}{llrr}
\toprule
Head & Mixing before readout & Width & Params \\
\midrule
\texttt{pooled}      & none                                   & --- & 264{,}449 \\
\texttt{axial}       & 1 row+column self-attention block      & 32  & 29{,}537 \\
\texttt{pairformer1} & 1 triangular pair block                & 64  & 231{,}233 \\
\texttt{pairformer2} & 2 triangular pair blocks               & 64  & 330{,}177 \\
\bottomrule
\end{tabular}
\end{table}

\paragraph{Training.}
Heads are trained with a masked per-residue binary cross-entropy loss,
\begin{equation}
\begin{aligned}
\mathcal{L} \;&=\; \frac{\sum_i m_i \, \ell_i}{\sum_i m_i},
\qquad
\ell_i \;=\; \mathrm{BCE}_{w,\gamma}\big(\hat{y}_i, y_i\big), \\[2pt]
\mathrm{BCE}_{w,\gamma}\big(\hat{y}, y\big) \;&=\; -\Big[\, w \, (1-\hat{y})^{\gamma} \, y \log \hat{y}
\;+\; \hat{y}^{\gamma} \, (1-y) \log (1-\hat{y}) \,\Big].
\end{aligned}
\label{eq:loss}
\end{equation}
Here $y_i$ is the residue's label, $\hat{y}_i$ the head's predicted probability, $m_i \in \{0,1\}$
the mask of Section~\ref{sec:dataset_construction}, $w$ the positive-class weight and $\gamma$ the
focal exponent~\citep{lin2017focal}; $\mathrm{BCE}_{1,0}$ is plain unweighted binary
cross-entropy. For the initial version, we did not mask any pockets during
training, but in the future we could use it to mask low-confidence pockets. Our reported models
use $w = 1$ and $\gamma = 0$, as selected on the validation set.

The trunk is frozen throughout, so only head parameters are updated and each
protein is featurized exactly once. Because lengths vary there is nothing to
stack into a batch: we take one protein per forward pass and accumulate gradients
over 8 proteins per optimizer step. We train for 20 epochs with
Adam~\citep{kingma2015adam} at a
learning rate of $2\times10^{-3}$ (40 epochs for the \texttt{single} heads,
which converge later), evaluate every 2 epochs, and select the
checkpoint with the best per-target validation \auprc{}, which is then scored once
on the held-out test split. We mask ``seen'' pockets in the validation set when computing metrics.

\paragraph{Clustering decoder.}
To obtain pocket centroids we group the residues the head is confident about by position and take
each group as one pocket. Residues scoring at or above a threshold $\tau$ seed a weighted mean
shift over their C$\alpha$ coordinates, each residue weighted by its predicted probability; modes
closer than half the bandwidth are merged, groups smaller than $m$ residues are discarded as
noise, and each surviving group is reported at the probability-weighted centroid of its residues'
heavy atoms, ranked by summed residue score. The three free parameters --- $\tau$, the bandwidth,
and $m$ --- are chosen by a grid search on validation scored by DCC top-$N$ at 4\,\AA{} and frozen before
test is read; the selected setting is $\tau = 0.3$, bandwidth 14\,\AA{}, $m = 3$. Two further
constants are set rather than tuned: convergence is declared at 0.05\,\AA{} or 100 iterations, and a
protein whose residues all fall below $\tau$ falls back to its ten highest-scoring residues so
that every target receives a prediction. Clustering per-residue predictions into discrete pockets
is challenging~\citep{fan2025unisite}, and we plan to explore different clustering approaches in
future iterations of \method{} work.

\paragraph{Re-ranking decoder.}
Some cavity detectors have high recall and poor ranking. DeepPocket \citep{aggarwal2022deeppocket} exploits this by keeping fpocket's proposals and re-ranking them with a learned 3D-CNN scored on a
grid around each cavity centroid. We follow the same division of labor with the
per-residue head in place of the CNN: each cavity is scored by aggregating the
head's probabilities over the residues lining it, and the cavities are re-ordered
by that aggregate. The lining is either the detector's own residue list or every
residue with a heavy atom within a radius of the cavity, and the aggregate is a
mean, a max, a sum, or the mean of the $k$ highest --- all chosen on validation. We apply this to all five detectors --- fpocket, P2Rank \citep{krivak2018p2rank},
Lacuna, and VN-EGNN \citep{sestak2024vnegnn} as published and retrained. We also apply this to
\emph{union}, the pool of every detector's cavities (other than zPocket), deduplicated at 2\,\AA{} so that two centroids on one cavity cannot spend two ranks. Since \emph{union} has no ranking of its own, it only appears re-ranked by zPocket. In the ``hybrid'' decoder, the \method{} cluster centroids are added to the union list and re-ranked using zPocket. The re-ranking hyper-parameters selected on the validation set are reported in Table~\ref{tab:decoders}.

\begin{table}[t]
\centering
\small
\setlength{\tabcolsep}{6pt}
\caption{Re-ranking configurations selected on validation. The grid spans seven aggregates
(mean, max, sum, and the mean of the $k$ highest for $k \in \{4, 8, 16, 32\}$) and four linings (the
cavity's stored residue list, or a 5, 8 or 10\,\AA{} heavy-atom shell around its centroid), giving 28 configurations per detector. \emph{stored list} is the detector's own lining where it
publishes one --- fpocket's alpha-sphere contacts, P2Rank's surface-point residues, Lacuna's
reported residues --- and, for VN-EGNN, which emits points rather than cavities, an 8\,\AA{} shell
computed around each virtual-node centroid at detection time. Configurations are determined separately for each trunk's head; the main text uses the ESMFold2 configurations.}
\label{tab:decoders}
\begin{tabular}{lllll}
\toprule
& \multicolumn{2}{c}{ESMFold2} & \multicolumn{2}{c}{Protenix} \\
\cmidrule(lr){2-3}\cmidrule(lr){4-5}
Proposals & Aggregate & Lining & Aggregate & Lining \\
\midrule
fpocket           & mean of top 16 & 5\,\AA{} shell  & mean of top 32 & 8\,\AA{} shell \\
P2Rank            & mean           & 8\,\AA{} shell  & mean           & 8\,\AA{} shell \\
Lacuna            & mean           & 8\,\AA{} shell  & mean           & 8\,\AA{} shell \\
Lacuna (SF-seq)   & mean           & stored list     & mean           & stored list \\
VN-EGNN (PW)      & mean           & 8\,\AA{} shell  & mean of top 16 & 10\,\AA{} shell \\
union             & mean           & stored list     & mean           & stored list \\
\bottomrule
\end{tabular}
\end{table}

\subsection{Baselines}
\label{app:baselines}

\paragraph{VN-EGNN.}
\label{app:vnegnn-unk}

We compare against both the released VN-EGNN checkpoint and a retrained version; however these two versions are run on different code paths and so are not directly comparable. The released checkpoint is the authors' own, but it runs on an older version of the code that only supported one site per protein by construction. The newly released training code supports multiple sites per protein, so we use that version to retrain as the PocketWatch dataset carries multiple sites per protein. The new code path further adds a residue-type one-hot to the graph representation in addition to the ESM-2 embeddings, so the model features are also different from the released version (1301 input dimension against 1280), which may also contribute to the performance difference. We trained for 300 epochs over the training proteins which survive featurization (32 out of 11{,}016 were dropped), keeping the checkpoint with the best validation DCC (checkpoint 269 on ESMFold2 folds, 274 on Protenix folds, from two otherwise identical runs). While the repo default is 1500 epochs, we found that the validation DCC had plateaued by 100 epochs and so we stopped early. We trained a single model using seed 42.

One forward pass produces both of VN-EGNN's outputs: the ranked centroid proposals used for the pocket-centroid tables and the per-residue logits used for the per-residue comparison. The pocket centroid proposals are where VN-EGNN's virtual nodes settle, and they are ranked by the confidence head. To get per-residue scores, we read the logits from the segmentation head which produces a per-residue logit that is trained with a Dice loss. To enable consistent comparison against other methods, we dropped the 4 proteins from our test set that contained \texttt{UNK} residues as these were not supported by either the released or re-trained version of VN-EGNN; however it may be possible to patch this issue in future iterations.

\paragraph{UniSite-1D.} The released repository ships inference and evaluation only, so we supply the optimization loop, the loss weighting and a dataset that serves our labels as set-prediction targets, adopting the Mask2Former conventions (classification 2, mask 5, Dice
5). Our retrained model is therefore UniSite's architecture and loss under hyperparameters we chose, not a reproduction of the authors' recipe, and should be read as a lower bound on what retraining this architecture on PocketWatch can reach. We train from scratch using batch size 8, for 30 epochs over the entire training set, selecting on validation per-residue
\auprc{}; we report the mean of three seeds, whose spread is 0.0025 SD on per-target \auprc{}.

UniSite-1D emits a set of scored masks, so for the per-residue tables we take, for each residue, the maximum over queries of the query's objectness times its mask probability there. Ranking by mask probability produces worse results (0.567 to 0.296), because many queries place confident mass on the wrong region and are separated only by their objectness scores. We use the model's continuous mask probabilities rather than the binary masks its post-processing returns. We use this same method to get per-residue scores for both the published and re-trained variants of the model. We note that the frozen model is partly leaked: 216 of our test proteins (13.8\%) appear in UniSite-DS, but we report it on the full test set regardless.

\paragraph{Lacuna.} We used \texttt{lacuna-pockets} 1.2.0, invoked per fold as
\texttt{lacuna discover} with \texttt{-{}-conformers 20}, reading the same
ESMFold2-predicted fold that the other baselines read. ESMFold2 folds carry no per-atom occupancy,
which Lacuna requires, so the copy of each fold Lacuna reads has occupancy set to 1.0. We report two variants: the
default detector and ranker (\texttt{-{}-detector alpha}, \texttt{-{}-rank-by
learned}), and surface fusion (\texttt{-{}-detector surface-fusion}), which pools
the geometric and learned-surface detectors and selects its own matching ranker
(\texttt{learned-fused}); its sequence pass requires the optional \texttt{plm}
extra and ESM-2 650M. Lacuna truncates its report to the ten best pockets by
default, which would cap recall and every ceiling we report, so we removed the cap by raising it to 1,000. The most any test protein produced
was 243 cavities under surface fusion and 232 under the default detector (171 and 162 on Protenix folds).

\subsection{Evaluation}
\label{app:metrics}

\paragraph{Residue-level metrics.}
Pocket finding is a heavily imbalanced per-residue retrieval problem, so we report threshold-free ranking metrics and their exact random expectations. Our headline metric is
\textbf{per-target \auprc{}}: average precision computed within each protein and
then averaged over proteins, which prevents large proteins from dominating. We
report \textbf{AUROC} for comparability with prior work and
\textbf{R-precision} (precision among the top-$k$ residues, where $k$ is that
target's number of pocket residues). Because a protein's
pockets are pooled into a single ranking by any per-target metric, we
additionally evaluate each pocket \textbf{in isolation}: its residues are the
positives, the chain's non-pocket residues the negatives, and every other pocket
is masked out. We then average over pockets, assigning equal weight. Every metric is computed only over unmasked residues.

\paragraph{Pocket-centroid metrics.}
A method emits a ranked list of pocket centroids, scored by distance to real
ligands. \textbf{DCC} is the distance from a predicted centroid to a site's
ground-truth ligand centroid, and a prediction succeeds if it is within 4\,\AA{}, a common
criterion in this literature. Following \citet{utges2024comparative}, we report the top-$N{+}2$ recall metric: we cut the predicted centroid list based on the protein's own site count plus two (DCC$_{N+2}$) and report the fraction of ground-truth sites found, testing each site against every kept prediction independently. A budgeted recall says nothing about what a method emits outside the budget, so we also report \textbf{average precision} (AP) over
the whole ranked list, which charges for every false positive ranked above a hit
and requires one-to-one matching (a centroid between two sites is one true
positive, not two), together with pooled precision and the mean list length
$\bar{k}$. Finally we report the \textbf{ceiling}: the same DCC rate with no
budget at all. The ceiling helps separate a detection failure from a ranking failure.

\subsection{Additional co-folding trunks}
\label{app:trunks}
Section~\ref{sec:results-trunks} compares \method{} heads on seven frozen trunks. This section gives
how the five added trunks were featurized and trained; Protenix-v2 and ESMFold2 are featurized as in Appendix~\ref{app:model}.

\paragraph{Checkpoints.}
Table~\ref{tab:trunk-settings} lists the checkpoint and inference settings for each added trunk.
We pin checkpoints whose experimental training structures were all deposited before 2021-09-30,
the start of the PocketWatch validation period (Section~\ref{sec:dataset_construction}), so no
trunk has seen a validation or test pocket; later releases of OpenFold3 and some Protenix
checkpoints use later cutoffs and were not used. For RF3 we use the September 2021 preprint
checkpoint rather than the package's default, which is trained on structures deposited before
January 2024 and so has seen the validation period. Undated distillation sets of predicted structures are
not covered by this check.

\begin{table}[!ht]
\centering
\small
\setlength{\tabcolsep}{2.5pt}
\caption{Inference settings for the added trunks. \emph{Cutoff} is the latest deposition date of
the checkpoint's experimental training structures. \emph{$z$ read from} is where the pair
representation is captured: in every case after the last recycle and before the diffusion module.
$^{\dagger}$Structures released before September 2021, as stated in the RF3
preprint~\citep{corley2025rf3}; the exact day is not given.}
\label{tab:trunk-settings}
\begin{tabular}{llccl}
\toprule
Trunk & Checkpoint & Cutoff & Recycles & $z$ read from \\
\midrule
Protenix-v1       & \texttt{base\_default\_v1.0.0} & 2021-09-30 & 10 & distogram-head input \\
OpenFold3-preview & \texttt{of3-p2-155k}           & 2021-09-30 & 3  & latent pair output \\
Boltz-1           & Boltz-1                        & 2021-09-30 & 3  & trunk embedding \\
Chai-1            & chai-lab 0.6.1                 & 2021-01-12 & 3  & trunk output \\
RF3               & \texttt{09\_21\_preprint}       & Sep.\ 2021$^{\dagger}$ & 10 & recycler output \\
\bottomrule
\end{tabular}
\end{table}

\paragraph{Featurization.}
Each trunk folds the same proteins as the main-text trunks, with MSAs read from one shared
per-sequence cache, no templates and a fixed seed, and with the confidence heads switched off where
the model allows. The heads read only $z$, so no structure-reading baseline is rerun on these
trunks.
Protenix-v1's inference-time dropout is disabled, and Chai-1 also reads ESM2-3B language-model
embeddings, as in its default configuration. As for the main-text trunks, $z$ is stored in half
precision. Non-standard residues (selenocysteine, pyrrolysine and ambiguous codes) are input as
unknown (X); RF3 represents such a residue with one token per atom, and the head's loader averages
those tokens back to one per residue (42 of the 14{,}135 proteins). Protenix-v1, OpenFold3-preview
and Boltz-1 were folded mostly on RTX PRO 6000 GPUs, on which Protenix-v1 is not bit-reproducible
above roughly 800 residues (run-to-run correlation of $z$ $\geq 0.997$); Chai-1 and RF3 were folded
mostly on H100 and A100 GPUs.

\paragraph{Heads and training.}
We train only \texttt{pooled} and \texttt{pairformer2}, the simplest and the best $z$ head, with
the recipe of Appendix~\ref{app:model} unchanged, on one training seed. Because \texttt{pooled} reads
$z$ at full width, its size scales with $c_z$: 165{,}377 parameters on the 128-wide trunks
against 264{,}449 on the 256-wide ones (Table~\ref{tab:heads}). \texttt{pairformer2}
projects $z$ to width 64 first, so it is nearly the same size on every trunk (321{,}985 against
330{,}177), which makes it the cleaner trunk comparison. Head capacity alone does not explain the
\texttt{pooled} ordering, since 128-wide Protenix-v1 beats 256-wide ESMFold2. The selected epochs
are 17 (Protenix-v1), 13 (OpenFold3-preview), 7 (Boltz-1), 15 (Chai-1) and 15 (RF3) for
\texttt{pairformer2}, and 17, 17, 15, 19 and 15 for \texttt{pooled}; Chai-1's \texttt{pooled}, like
the Protenix-v2 and ESMFold2 \texttt{pooled} heads, peaks at the last evaluated epoch. Boltz-1's \texttt{pairformer2} peaks early: its validation \auprc{}
plateaus after epoch 7 while training loss keeps falling, the same mild overfitting as Protenix-v2's
\texttt{pairformer2} (best epoch 11).

\paragraph{Scoring.}
Heads are scored exactly as in Table~\ref{tab:main}: the same targets, masks and site policy.
Intervals are percentile bootstraps over targets (4{,}000 resamples), and paired differences
resample the per-target difference between two heads of the same type.

\newpage

\section{Supplementary results}
\label{app:results}

\subsection{Additional per-residue results}
\label{app:perresidue-results}
\begin{figure}[!ht]
\centering
\includegraphics[width=\textwidth]{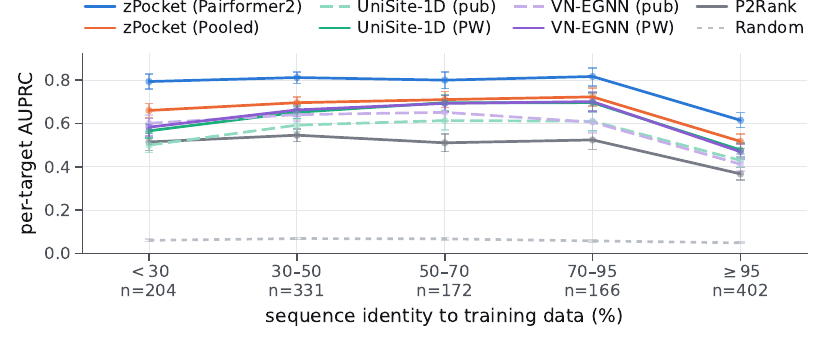}
\caption{Per-target \auprc{} against sequence identity to anything the model has already seen ---
the trunk's pretraining sequences plus the PocketWatch training and validation sequences.
Performance is roughly flat from \emph{novel} through \emph{close} and falls only in the
\emph{near-identical} ($\geq$95\%) bin. We suspect that this fall is an artifact of the seen-pocket mask rather than a generalization effect: a protein with pockets deposited before the cutoff has
most of its familiar pockets masked, leaving the harder remainder to be scored. Bin sizes are printed on the axis and bars are bootstrap 95\% intervals. \method{} outperforms all baselines.}
\label{fig:sequence}
\end{figure}

Table~\ref{tab:full} extends the full per-residue comparison in
Table~\ref{tab:main} with a trunk comparison: every \method{} head on the ESMFold2 trunk alongside the same head on the Protenix trunk, with P2Rank on each trunk's own predicted folds as a reference. The learned baselines are left to Table~\ref{tab:main}. Figure~\ref{fig:sequence} breaks per-target \auprc{} down by sequence identity to anything the model has already seen.

\begin{table}[t]
\centering
\small
\setlength{\tabcolsep}{1.8pt}
\caption{\textbf{Protenix against ESMFold2 as a frozen trunk.} Per-residue results on
PocketWatch for every \method{} head trained on each of the two trunks, scored on the 1{,}562
pockets of 1{,}275 targets in the effective test set. Protenix leads on every matched head, but the margin is small compared to the choice of
head architecture. The ranking across \method{} head architectures is the same for both trunks. The \texttt{single} rows are not a
matched pair: Protenix uses the trunk single representation $s$, while ESMFold2 uses the final-layer ESMC-6B embedding. P2Rank, run on each co-folding model's own predicted folds, is included as a structure-dependent reference point on the same folds the heads are compared over. Columns: 
Rp = R-precision. Per-target metrics are computed once per target from that target's residue
ranking, then averaged over targets with equal weight. Per-pocket metrics are computed once per
pocket, with positives from other pockets masked out, and averaged over pockets with equal
weight. The last two columns split AUPRC$_{p}$ by the number of scored pockets for each protein.
Bold = best per column.}
\label{tab:full}
\begin{tabular}{llcccccccc}
\toprule
& & \multicolumn{3}{c}{per target} & \multicolumn{3}{c}{per pocket} & \multicolumn{2}{c}{AUPRC$_{p}$ by multiplicity} \\
\cmidrule(lr){3-5}\cmidrule(lr){6-8}\cmidrule(lr){9-10}
Trunk & Head & AUPRC$_{t}$ & AUROC$_{t}$ & Rp$_{t}$ & AUPRC$_{p}$ & AUROC$_{p}$ & Rp$_{p}$ & 1 pocket & $>$1 \\
\midrule
Protenix & single ($s$) & 0.474 & 0.874 & 0.444 & 0.442 & 0.863 & 0.413 & 0.468 & 0.390 \\
 & \texttt{pooled} & 0.669 & 0.938 & 0.614 & 0.633 & 0.927 & 0.579 & 0.667 & 0.560 \\
 & \texttt{axial} & 0.724 & 0.950 & 0.658 & 0.683 & 0.941 & 0.620 & 0.726 & 0.594 \\
 & \texttt{pairformer1} & 0.746 & 0.956 & 0.680 & 0.702 & 0.946 & 0.637 & 0.750 & 0.602 \\
 & \texttt{pairformer2} & \textbf{0.757} & \textbf{0.958} & \textbf{0.692} & \textbf{0.714} & \textbf{0.948} & \textbf{0.650} & \textbf{0.761} & \textbf{0.617} \\
\midrule
ESMFold2 & single (ESMC) & 0.611 & 0.918 & 0.564 & 0.574 & 0.906 & 0.528 & 0.610 & 0.499 \\
 & \texttt{pooled} & 0.639 & 0.924 & 0.589 & 0.600 & 0.913 & 0.551 & 0.640 & 0.516 \\
 & \texttt{axial} & 0.716 & 0.945 & 0.652 & 0.675 & 0.934 & 0.613 & 0.719 & 0.584 \\
 & \texttt{pairformer1} & 0.737 & 0.950 & 0.677 & 0.695 & 0.941 & 0.637 & 0.741 & 0.600 \\
 & \texttt{pairformer2} & 0.745 & 0.951 & 0.682 & 0.702 & 0.941 & 0.639 & 0.751 & 0.599 \\
\midrule
\multicolumn{2}{l}{P2Rank (Protenix folds)} & 0.503 & 0.869 & 0.473 & 0.479 & 0.864 & 0.447 & 0.495 & 0.444 \\
\multicolumn{2}{l}{P2Rank (ESMFold2 folds)} & 0.477 & 0.859 & 0.449 & 0.455 & 0.855 & 0.425 & 0.466 & 0.430 \\
\multicolumn{2}{l}{random} & 0.049 & 0.500 & 0.049 & 0.051 & 0.500 & 0.051 & 0.053 & 0.048 \\
\bottomrule
\end{tabular}
\end{table}

\subsection{Additional results for additional co-folding trunks}
\label{app:additional-trunk-results}
Table~\ref{tab:trunks-full} gives the full comparison between different co-folding trunks.
\begin{table}[!ht]
\centering
\small
\setlength{\tabcolsep}{3pt}
\caption{Per-target results for every trunk and head, as in Table~\ref{tab:trunks} with AUROC,
R-precision (Rp) and paired differences in \auprc{} against the two main-text trunks (head minus
reference, same head type, 95\% CI over 4000 bootstrap samples). Each row is a single training run.}
\label{tab:trunks-full}
\begin{tabular}{lccccc}
\toprule
Trunk & AUPRC$_{t}$ & AUROC$_{t}$ & Rp$_{t}$ & $\Delta$ vs ESMFold2 & $\Delta$ vs Protenix-v2 \\
\midrule
\multicolumn{6}{l}{\texttt{pairformer2}} \\
\quad Protenix-v2        & 0.757 & 0.958 & 0.692 & $+0.011$ $[+0.003, +0.020]$ & --- \\
\quad Chai-1             & 0.746 & 0.955 & 0.680 & $+0.000$ $[-0.008, +0.008]$ & $-0.011$ $[-0.020, -0.003]$ \\
\quad ESMFold2           & 0.745 & 0.951 & 0.682 & --- & $-0.011$ $[-0.020, -0.003]$ \\
\quad Protenix-v1        & 0.742 & 0.953 & 0.677 & $-0.004$ $[-0.012, +0.004]$ & $-0.015$ $[-0.023, -0.007]$ \\
\quad OpenFold3-preview  & 0.742 & 0.952 & 0.680 & $-0.003$ $[-0.010, +0.004]$ & $-0.015$ $[-0.022, -0.007]$ \\
\quad RF3                & 0.741 & 0.950 & 0.676 & $-0.004$ $[-0.012, +0.003]$ & $-0.016$ $[-0.024, -0.008]$ \\
\quad Boltz-1            & 0.733 & 0.949 & 0.665 & $-0.012$ $[-0.021, -0.004]$ & $-0.024$ $[-0.032, -0.015]$ \\
\midrule
\multicolumn{6}{l}{\texttt{pooled}} \\
\quad Protenix-v2        & 0.669 & 0.938 & 0.614 & $+0.030$ $[+0.024, +0.035]$ & --- \\
\quad Chai-1             & 0.668 & 0.935 & 0.612 & $+0.029$ $[+0.023, +0.034]$ & $-0.001$ $[-0.006, +0.004]$ \\
\quad ESMFold2           & 0.639 & 0.924 & 0.589 & --- & $-0.030$ $[-0.035, -0.024]$ \\
\quad Protenix-v1        & 0.650 & 0.931 & 0.595 & $+0.011$ $[+0.005, +0.017]$ & $-0.019$ $[-0.023, -0.014]$ \\
\quad OpenFold3-preview  & 0.619 & 0.922 & 0.566 & $-0.020$ $[-0.026, -0.014]$ & $-0.050$ $[-0.055, -0.044]$ \\
\quad RF3                & 0.631 & 0.927 & 0.576 & $-0.008$ $[-0.014, -0.002]$ & $-0.038$ $[-0.043, -0.033]$ \\
\quad Boltz-1            & 0.638 & 0.929 & 0.585 & $-0.001$ $[-0.007, +0.005]$ & $-0.031$ $[-0.036, -0.026]$ \\
\bottomrule
\end{tabular}
\end{table}

\subsection{Additional pocket centroid results}
\label{app:centroid-results}
\begin{figure}[!ht]
\centering
\includegraphics[width=0.95\textwidth]{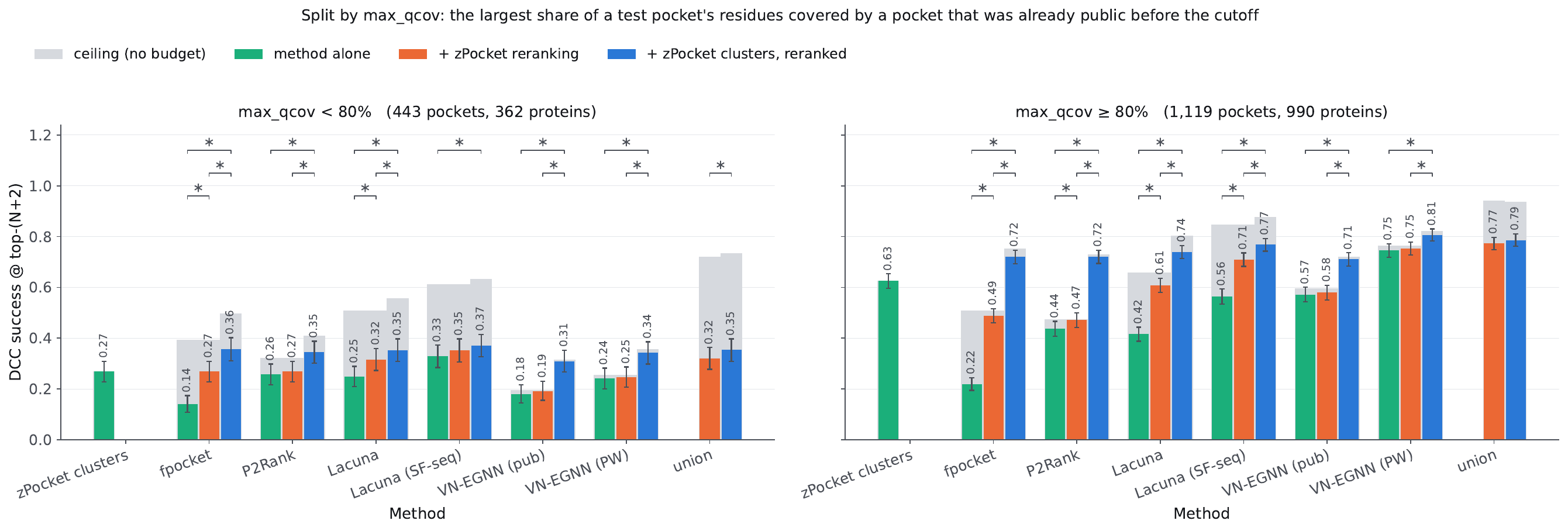}
\caption{\textbf{Pocket centroid prediction split by pocket familiarity.} The comparison of
Figure~\ref{fig:treatments}, DCC$_{N+2}$ at 4\,\AA{}, computed separately on unfamiliar pockets
(\qcov{} below 80\%: 443 pockets of 362 proteins) and familiar ones (\qcov{} at or above 80\%:
1{,}119 pockets of 990 proteins). Colors, gray ceilings and error bars are
as in Figure~\ref{fig:treatments}: green is each method's own proposals under its own ranking,
orange re-ranks those proposals with the \method{} head, blue adds the \method{} cluster
proposals before re-ranking, an asterisk marks a paired 95\% bootstrap excluding zero, and the
intervals on the bars themselves are unpaired. Every method degrades sharply on unfamiliar
pockets, and the gap between each bar and its own ceiling widens, but adding the \method{}
proposals and re-ranking still lifts DCC$_{N+2}$ significantly for most methods in both similarity bins.}
\label{fig:treatments-qcov}
\end{figure}

Table~\ref{tab:discrete} gives the numbers behind Figure~\ref{fig:treatments} --- DCC at both
budgets, the ceiling, average precision, precision and the emission rate --- for every proposal
and ranker combination. Figure~\ref{fig:treatments-qcov} repeats the same comparison split by
\qcov{} (Section~\ref{app:qcov}), the largest share of a test pocket's residues covered by a
pocket that was already public before the cutoff.

\begin{table}[t]
\centering
\small
\setlength{\tabcolsep}{1.5pt}
\caption{Discrete pocket prediction on PocketWatch with the ESMFold2 \texttt{pairformer2} head and every detector run on ESMFold2 folds: how close a ranked list of predicted centroids lands to the real ligand centroids, over the same 1{,}562 pockets of 1{,}275 targets. A prediction succeeds within 4~\AA{} of a ground-truth centroid. DCC $N$ keeps each protein's own pocket count as the budget and $N{+}2$ allows two more; \emph{ceiling} is the rate with no budget at all, which the proposals fix and no ranking can exceed; \emph{prec.} is the share of everything emitted that claimed a pocket, and $\bar{k}$ how many centroids were emitted per protein. Every decoder is the winner of a validation sweep restricted to ranking by a single scorer, so no row mixes the head's score with a detector's own. \emph{union} pools every detector's cavities, folded most-selective first and deduplicated, so it has no ranking of its own to report -- only a pool for the head to rank. Bold = best per column.}
\label{tab:discrete}
\begin{tabular}{lllcccccc}
\toprule
& & & \multicolumn{2}{c}{DCC success} & & & & \\
\cmidrule(lr){4-5}
Section & Proposals & Ranker & $N$ & $N{+}2$ & ceil. & AP & prec. & $\bar{k}$ \\
\midrule
Base methods & fpocket & fpocket & 0.139 & 0.197 & 0.476 & 0.195 & 0.014 & 41.3 \\
 & P2Rank & P2Rank & 0.324 & 0.386 & 0.431 & 0.371 & 0.044 & 11.8 \\
 & Lacuna & Lacuna & 0.209 & 0.369 & 0.616 & 0.310 & 0.013 & 57.5 \\
 & Lacuna (SF-seq) & Lacuna (SF-seq) & 0.273 & 0.497 & 0.780 & 0.417 & 0.015 & 64.9 \\
 & VN-EGNN (pub) & VN-EGNN (pub) & 0.417 & 0.461 & 0.483 & 0.482 & 0.124 & 4.8 \\
 & VN-EGNN (PW) & VN-EGNN (PW) & 0.533 & 0.602 & 0.620 & 0.606 & 0.179 & 4.2 \\
 & zPocket clusters & zPocket & 0.515 & 0.524 & 0.524 & 0.570 & \textbf{0.599} & 1.1 \\
\midrule
Rescoring with zPocket & fpocket & zPocket & 0.371 & 0.426 & 0.476 & 0.426 & 0.014 & 41.3 \\
 & P2Rank & zPocket & 0.386 & 0.414 & 0.431 & 0.420 & 0.044 & 11.8 \\
 & Lacuna & zPocket & 0.399 & 0.525 & 0.616 & 0.489 & 0.013 & 57.5 \\
 & Lacuna (SF-seq) & zPocket & 0.405 & 0.608 & 0.780 & 0.539 & 0.015 & 64.9 \\
 & VN-EGNN (pub) & zPocket & 0.440 & 0.470 & 0.483 & 0.498 & 0.124 & 4.8 \\
 & VN-EGNN (PW) & zPocket & 0.538 & 0.609 & 0.620 & 0.612 & 0.179 & 4.2 \\
 & union & zPocket & 0.497 & 0.645 & \textbf{0.879} & 0.627 & 0.009 & 114.4 \\
\midrule
Hybrid & zPocket + fpocket & zPocket & 0.534 & 0.617 & 0.680 & 0.621 & 0.020 & 42.3 \\
 & zPocket + P2Rank & zPocket & 0.542 & 0.614 & 0.639 & 0.615 & 0.061 & 12.7 \\
 & zPocket + Lacuna & zPocket & 0.537 & 0.629 & 0.734 & 0.631 & 0.015 & 58.4 \\
 & zPocket + Lacuna (SF-seq) & zPocket & 0.538 & 0.656 & 0.808 & 0.647 & 0.015 & 65.7 \\
 & zPocket + VN-EGNN (pub) & zPocket & 0.532 & 0.597 & 0.606 & 0.609 & 0.210 & 3.5 \\
 & zPocket + VN-EGNN (PW) & zPocket & \textbf{0.544} & \textbf{0.675} & 0.690 & 0.648 & 0.213 & 4.0 \\
 & zPocket + union & zPocket & 0.542 & 0.664 & \textbf{0.879} & \textbf{0.659} & 0.009 & 113.6 \\
\bottomrule
\end{tabular}
\end{table}

\subsection{How do predicted structures affect P2Rank?}
\label{app:apo-check}
Every number in this paper is computed on a predicted structure. It is possible that pocket prediction tools like P2Rank, which can work on either predicted or experimental structures, might perform differently depending on the input. To check this, we ran P2Rank on a set of experimental apo structures and compared the results to those obtained from the predicted folds.

For every test target with a deposited apo chain we selected one experimental structure and
ran the same P2Rank configuration on it. A chain was admitted only if it was ligand-free at
every scored pocket residue (no non-polymer component within 5\,\AA{}, ignoring
crystallization additives and anything under five heavy atoms), contained all annotated pockets, and came from a structure at 2.5\,\AA{} or better; among admissible chains we took the best resolution. We then ran P2Rank on this experimental apo structure and compared its results to those obtained from the predicted fold of the same target, only considering the canonical positions the experimental chain observes (sometimes the experimental chain is a subset of the full UniProt sequence). This procedure gave us an evaluation set with 433 targets
and 537 pockets (398 X-ray, 35 cryo-EM; median resolution 1.85\,\AA{}).

We found that P2Rank performs similarly on both structure sources. Per-target \auprc{} is $0.518$ on the experimental apo
structure against $0.505$ on the ESMFold2 fold of the same target, a paired difference of
$+0.013$ (95\% percentile bootstrap CI over targets, $[-0.003, +0.030]$); per pocket,
$0.489$ against $0.473$ ($+0.016$, $[+0.000, +0.032]$. The gap is larger than on the
Protenix fold, where the same comparison gives $0.514$ and a paired difference of $+0.004$
($[-0.013, +0.022]$), and it comes almost entirely from the 49 targets whose apo chain observes
fewer than half of the scored residues. The intervals exclude an advantage for the experimental
structure larger than about $0.03$ \auprc{}, an order of magnitude smaller than \method{}'s margin
over P2Rank in Table~\ref{tab:full}. We therefore read the fold-based numbers throughout as a fair
proxy for what these methods would reach on experimental input.

\subsection{Crypticity versus predicted-fold accuracy}
\label{app:crypticity-results}
Since cryptic sites might be difficult to predict from a single conformation of an unbound protein, we were eager to evaluate whether a method like \method{}, which does not depend on coordinates, would predict these pockets better than methods that do. In the course of our
investigation, we found that the predicted folds often come close to the ground-truth pocket even for sites labeled cryptic per our definition,
and that some sites labeled not cryptic had predicted folds more than 2\,\AA{} away, so we decided to stratify our results by fold accuracy as well as by the cryptic label.
We think this is a useful finding about the ``cryptic'' definition used by IsoDDE and CryptoBench: it is a property of the experimental structures, not of the predicted structure, and a predicted apo fold can reproduce a holo pocket even if the model has never seen that holo conformation for this particular protein, perhaps by learning it from homologues.

\begin{figure}[!ht]
\centering
\includegraphics[width=\textwidth]{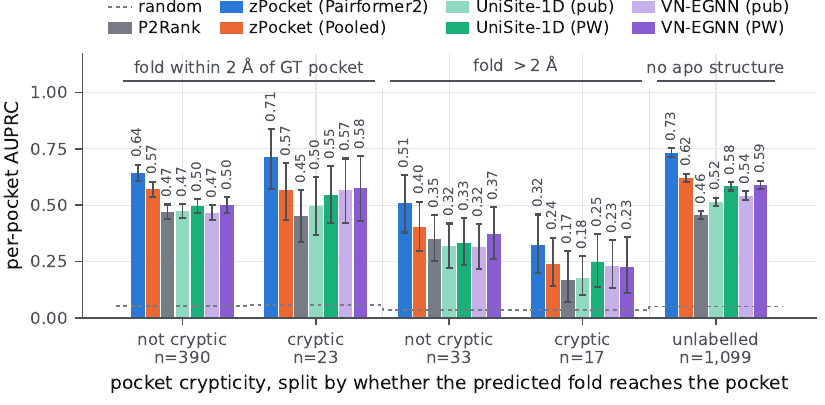}
\caption{Per-pocket \auprc{} by crypticity label, crossed with whether the ESMFold2 predicted
structure comes within 2\,\AA{} of the ground-truth pocket. \method{} heads against P2Rank and both baselines, each
as published (\emph{pub}) and retrained on PocketWatch (\emph{PW}). Bucket sizes are printed beneath each group; bars are bootstrap 95\%
intervals and the dashed line is the random expectation, computed per bucket from its own positive rate.}
\label{fig:cryptic}
\end{figure}

Fold accuracy is more predictive of binding prediction performance than crypticity: when the
predicted fold comes within 2\,\AA{} of the ground-truth pocket, both \method{} and P2Rank score
higher regardless of crypticity label, and when it does not, every method degrades
(Fig.~\ref{fig:cryptic}). Within the ``fold within 2\,\AA{} of ground-truth pocket'' bucket,
P2Rank shows a small dip for cryptic pockets (0.451 against 0.472), while \method{} scores higher on
them (0.713 against 0.644), but three of the five buckets hold only 33, 23 and 17 pockets, so it is
hard to draw conclusions from any of them.
Keeping in mind these small numbers, we note that \method{} still outperforms
baselines across every stratification. 

\subsection{Calibration and the main-text case-study figure}
\label{app:calibration-figure}

\begin{figure}[!ht]
\centering
\includegraphics[width=160.6pt]{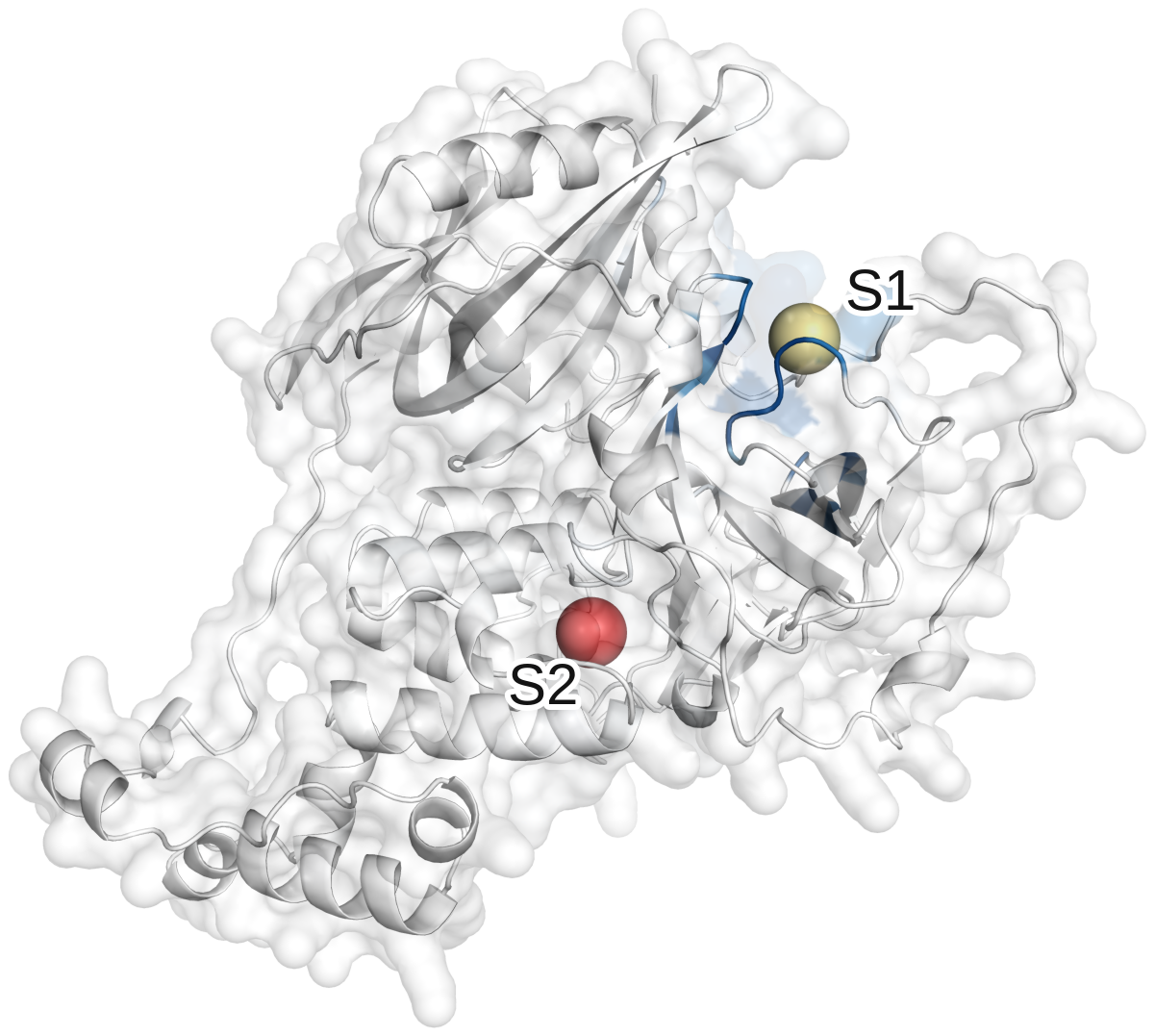}\par\vspace{4pt}
\includegraphics[width=0.5\textwidth]{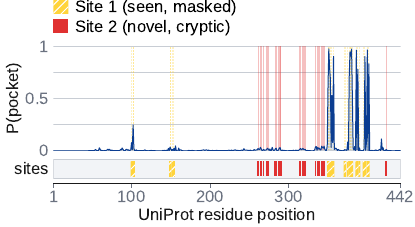}
\caption{\textbf{The main-text case study rendered from the reported model.} The same
target and annotated sites as Figure~\ref{fig:overview}(B), scored by the plain unweighted BCE
head ($w=1$, $\gamma=0$) used throughout the paper. Site coloring and seen-pocket masking follow Figure~\ref{fig:overview}(B).
The cartoon's color ramp spans scores 0--1 here and 0--0.4 in Figure~\ref{fig:overview}(B), where the
weighted-focal head's scores are compressed.}
\label{fig:calibration}
\end{figure}

Panel (B) of Figure~\ref{fig:overview} is rendered from a head trained with a weighted focal
loss ($w=5$, $\gamma=2$) rather than the plain unweighted BCE ($w=1$, $\gamma=0$) of
Eq.~\ref{eq:loss} that produces every quantitative result in this paper. The two heads are
otherwise identical. We show this variant because its higher positive weight and flatter calibration make $S_2$
easier to visualize, and we report the
plain-BCE head because it scores higher on the validation set ($0.766$ against $0.757$
\auprc{} on ESMFold2; $0.774$ against $0.766$ on Protenix). In fact, even though the weighted-focal head makes $S_2$ easier to see, the
plain-BCE head actually ranks the residues of $S_2$ higher, with a per-pocket
\auprc{} of $0.403$ compared to $0.247$ for the weighted-focal head. So the main text does not flatter the
case study by showing the weighted-focal variant. We show the same target scored by the
reported model in Figure~\ref{fig:calibration}.

\subsection{Fragment screens on the Protenix-v2 trunk}
\label{app:fragsoak-protenix}
\begin{figure}[!ht]
\centering
\includegraphics[width=\textwidth]{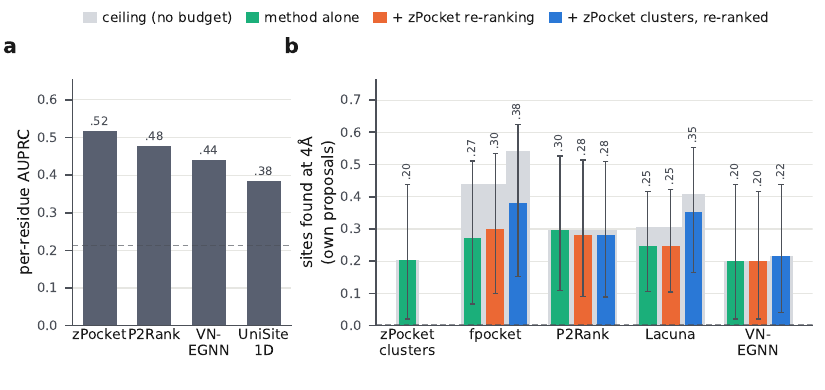}
\caption{\textbf{Figure~\ref{fig:fragsoak} on the \textbf{Protenix-v2} trunk.} Panels (a) and (b) of
Figure~\ref{fig:fragsoak}, with the Protenix-v2 \texttt{pairformer2} head and every detector run on
Protenix-v2 folds, over the same $35$ sites of $12$ proteins. Colors, ceilings, budget, success
radius and error bars as in Figure~\ref{fig:fragsoak}.}
\label{fig:fragsoak-protenix}
\end{figure}

Figure~\ref{fig:fragsoak} reports the fragment screens on ESMFold2 folds, the trunk the main text
uses throughout. Figure~\ref{fig:fragsoak-protenix} repeats its panels (a) and (b) on Protenix-v2,
over the same $35$ sites of $12$ proteins: the label-transfer check is required to pass on both
trunks, so the two are scored on identical labels. The worked examples of Figure~\ref{fig:fragsoak}c
are not repeated, being drawn on ESMFold2 folds.

Per-residue scores barely move between the trunks --- $0.518$ \auprc{} against $0.494$ --- but the
decoded centroids do, and the ceilings say why, though which detector the \method{} clusters help
differs by trunk. Pooling them into fpocket's or Lacuna's cavities adds nothing on ESMFold2, leaving
both ceilings where they were, while on Protenix-v2 it lifts fpocket's from $0.438$ to $0.542$ and
Lacuna's from $0.306$ to $0.410$, and both arms gain with them ($0.273$ to $0.382$ and $0.247$ to
$0.352$): the clusters propose sites those two detectors miss on this trunk and not on the other.
P2Rank runs the other way, its ceiling rising from $0.368$ to $0.385$ on ESMFold2 and its arm with it
($0.285$ to $0.301$), while on Protenix-v2 it sits on its ceiling ($0.297$). The published VN-EGNN
gains the same small margin on either trunk ($0.199$ to $0.216$), and the \method{} clustering decoder
sits on its own ceiling throughout ($0.100$ on ESMFold2, $0.204$ on Protenix-v2).

\clearpage

\section{Sourcing and curation of the fragment-soak test set}
\label{app:fragment-screening}

Fragment screens were taken from Diamond Light Source's XChem programme through Fragalysis,
downloaded on 2026-09-20/21 from the legacy host (per-target archives) and from the production and
staging hosts of the current stack (an unauthenticated REST interface), giving $61$ distinct
campaigns; one of them, the EV-A71 2A protease campaign, is the first OpenBind release
\citep{openbind2026}. Campaigns were selected before any prediction was run, on two independent
leakage axes computed for all $61$. On the \emph{head} axis a campaign is discarded when its
accession sits in the head's training split, or when a training-split protein exceeds $50\%$
identity to the construct. The rest are retained, whether the accession is absent from the label
table, sits in its validation or test split, or has a training relative below that bar; the closest
training relative among the campaigns kept is $44\%$. On the \emph{trunk} axis we count PDB entities at $\geq
70\%$ identity released before the co-folding trunk's 2021-09-30 cutoff, and discard a campaign
carrying more than fifteen. Raw counts mislead in both directions, so the selection also recorded
the two quantities behind them. The fraction of a protein's entities that predate the cutoff
separates campaigns the trunk merely glimpsed from campaigns it knows as well as it ever will:
A71EV2A and SOS3ATA both look sparse at $6$ and $3$ pre-cutoff entities, but A71EV2A has $480$
entities in total while all three of SOS3ATA's predate the cutoff, so the second was dropped and
the first kept. The count of pre-cutoff relatives at $30\%$ identity separates a sparsely sampled
protein from a densely sampled fold: CAMK1DA has ten pre-cutoff relatives at $70\%$ identity but
$1{,}079$ at $30\%$, because it is a kinase, and a per-residue head reading a trunk representation
is likelier to transfer at fold level than an identity tier implies, so it too was dropped. The
screens excluded this way include the SARS-CoV-2 main protease and nsp13 campaigns, which are
trunk-saturated, and the human NUDT, BRD and PTP1B campaigns, whose accessions are in the head's
training split. Eleven campaigns survived.

The Fragalysis campaigns are extended with fragment screens deposited straight into the PDB as group
depositions, which the Fragalysis release does not carry. Every group deposition whose entries were
released after the trunk's 2021-09-30 cutoff was collected ($132$ campaigns) and screened on the same
two leakage axes, leaving $16$ clean on both. Nine of those sixteen are proteins the Fragalysis set
already holds, in one case through a close homolog: four are the Coxsackievirus A16 2A protease, a
homolog of the EV-A71 2A protease already scored, and five are repeat screens of nsp14, the
chikungunya nsp3 macrodomain, the plant thioesterase FatA and the \emph{Porphyromonas gingivalis}
dipeptidyl-peptidase. All nine are dropped so that no protein is scored twice. Of the seven
that remain, FabF is discarded on the head axis, its closest training relative being at $66\%$
identity, leaving six new campaigns: TRIM21 ($109$ entries), ARHGEF2 ($56$), the SARS-CoV-2
nucleoprotein ($51$), ORF9b ($51$), the de novo designed protein ABLE ($43$) and MYLIP ($29$).

A site can only be scored on a structure the folding model is confident in, so every construct in the
test set, from either source, is required to reach a mean pLDDT of $85$ on the fold it is scored on
--- near the top of the conventional \emph{confident} band ($70$--$90$) and just below the
\emph{very high} boundary at $90$. Two of the group depositions fall far short and are dropped without
being scored: ORF9b, whose MSA carries only $39$ sequences, reaches $41.7$, and MYLIP $69.5$. Every
remaining construct clears the gate, spanning $85.5$ to $97.5$.

Within a campaign, a hit is the soaked compound's own coordinate file rather than a \texttt{HETATM}
record, because the deposited complexes also carry acetate, DMSO and ethylene glycol that were never
in the library; a fragment is counted once per distinct canonical SMILES however many crystals it
appears in. Every bound structure was checked to sit already in its campaign's apo reference frame
(C$\alpha$ deviation below $1.5$\,\AA{}) and ligand coordinates were used as deposited. Where a
second protein chain is present it is either a crystallographic copy of the modeled chain or a
genuine partner, and the two demand opposite treatment: ligands on a copy are carried onto the
modeled chain by superposition, while ligands at a partner interface (nsp16/nsp10, NXF1/NXT1) are
dropped and flagged. The two cases are told apart by sequence identity to the modeled chain and
never by chain label, since a campaign's template chain letter frequently disagrees with its bound
files. One campaign needs a further distinction: the $925$ site observations released for A71EV2A
are $113$ fragment-screen events followed by $812$ lead-like compounds from the optimization phase
that followed, and treating all $925$ as fragments saturates the active site and inflates the
positive base rate to most of the accessible surface, so only the fragment-screen subset is used.
Sites are then single-linkage clusters of ligand centroids at $5$\,\AA{} in the apo frame, the rule
PocketWatch already uses. Across both sources this produces $155$ sites, of which $117$ are
masked by rules applied uniformly to every campaign: $83$ carry fewer than three distinct fragments,
$20$ are crystal-packing artifacts (a median symmetry-mate contact fraction $\geq 0.25$ over the
site's fragments), $10$ have fewer than three lining residues that map onto the predicted fold, $2$
sit at a partner interface, and $2$ reach a predicted fold only through a lining-C$\alpha$ fit above
$2$\,\AA{}. A site is masked as well when its lining overlaps a pocket the head saw in training,
though with no retained accession in the training split that rule now removes none: a validation or
test pocket is a pocket the head was never shown, and is scored. The
transfer check is the one rule that depends on which structure is being scored, so it is required to
pass on both trunks' folds and the two are then scored on identical labels; three NXT1-NXF2 sites
transfer cleanly onto the ESMFold2 fold but not onto the Protenix-v2 one, whose model of that
construct is $7.9$\,\AA{} from the crystal template against ESMFold2's $2.0$\,\AA{}. This leaves the
$35$ scored sites over $12$ proteins that the main text reports, $29$ of them from Fragalysis and $6$
from the group depositions, with one Fragalysis campaign retaining none.

\section{Pocket novelty filters}
\label{app:pocket-filters}

The proteome scan, and every count, threshold and candidate in this appendix and
Appendix~\ref{app:high-conf-pockets}, uses the Protenix-v2 \texttt{pairformer2} head and
Protenix-v2 predicted folds. It was run while the Protenix-v2 weights were openly licensed and has not
been repeated on ESMFold2.

\begin{table}[t]
\centering
\caption{Pocket filtering funnel over the reviewed human proteome, grouped by how
evidence of a known site is propagated to the query: not at all, by sequence homology to
Pfam family and clan relatives, or by structural-motif similarity irrespective of
homology. Geometric requirements are independent of all three. Counts are pockets
remaining after each successive filter; proteins are the distinct accessions those pockets
lie on. Filters are conjunctive, so the survivor count is independent of their order;
per-row removals are not.}
\label{tab:pocket-filters}
\begin{tabular}{lrrr}
\toprule
Filter & Pockets & Removed & Proteins \\
\midrule
All predicted pockets & 21,646 & --- & 17,527 \\
\midrule
\multicolumn{4}{l}{\emph{Direct: evidence on the query accession}} \\
\quad Known ligand site on this protein          & 18,908 & 2,738 & 15,254 \\
\quad UniProt ACT\_SITE, BINDING or SITE         & 15,557 & 3,351 & 12,468 \\
\quad Polymer interface in its own assemblies    & 12,480 & 3,077 &  9,995 \\
\midrule
\multicolumn{4}{l}{\emph{Homology-propagated: transferred from Pfam family and clan relatives}} \\
\quad Transferred ligand site                    &  9,038 & 3,442 &  7,268 \\
\quad Transferred active site                    &  7,044 & 1,994 &  5,709 \\
\quad Transferred interface (family only)        &  4,489 & 2,555 &  3,590 \\
\quad Catalytic site localized, if an enzyme     &  4,456 &    33 &  3,564 \\
\midrule
\multicolumn{4}{l}{\emph{Geometry}} \\
\quad 9--30 lining residues                      &  1,024 & 3,432 &  1,017 \\
\quad Burial $\geq$ 39.9 heavy atoms             &    765 &   259 &    758 \\
\quad Mean pLDDT $\geq$ 80                       &    612 &   153 &    606 \\
\quad Mean head confidence $\geq$ 0.70           &    306 &   306 &    303 \\
\midrule
\multicolumn{4}{l}{\emph{Motif-propagated: structural similarity, no homology required}} \\
\quad Motif holds no ligand elsewhere in the PDB & \textbf{155} & 151 & \textbf{154} \\
\bottomrule
\end{tabular}
\end{table}

\begin{table}[h]
\centering
\caption{The funnel with all interface evidence removed, for comparison with
Table~\ref{tab:pocket-filters}. Geometric and motif requirements are unchanged. The
surviving set is a strict superset of the 155 reported there.}
\label{tab:pocket-filters-nointerface}
\begin{tabular}{lrrr}
\toprule
Filter & Pockets & Removed & Proteins \\
\midrule
All predicted pockets & 21,646 & --- & 17,527 \\
\midrule
\multicolumn{4}{l}{\emph{Direct}} \\
\quad Known ligand site on this protein          & 18,908 & 2,738 & 15,254 \\
\quad UniProt functional site, interactions excluded & 15,602 & 3,306 & 12,501 \\
\midrule
\multicolumn{4}{l}{\emph{Homology-propagated}} \\
\quad Transferred ligand site                    & 11,297 & 4,305 &  9,110 \\
\quad Transferred active site                    &  8,956 & 2,341 &  7,268 \\
\quad Catalytic site localised, if an enzyme     &  8,914 &    42 &  7,235 \\
\midrule
\multicolumn{4}{l}{\emph{Geometry}} \\
\quad 9--30 lining residues                      &  2,381 & 6,533 &  2,366 \\
\quad Burial $\geq$ 39.9 heavy atoms             &  1,828 &   553 &  1,816 \\
\quad Mean pLDDT $\geq$ 80                       &  1,579 &   249 &  1,568 \\
\quad Mean head confidence $\geq$ 0.70           &    754 &   825 &    750 \\
\midrule
\multicolumn{4}{l}{\emph{Motif-propagated}} \\
\quad Motif holds no ligand elsewhere in the PDB & \textbf{359} & 395 & \textbf{358} \\
\bottomrule
\end{tabular}
\end{table}

\subsection{Why known sites are resolved by homology}

A filter that asks whether a pocket coincides with a known binding site on \emph{its own}
protein cannot distinguish two outcomes: the site was tested and is clear, or nothing was
ever known about that protein. Keyed on the query accession, our annotation reached only
2{,}444 of the 20{,}092 scanned proteins, so 87\% of pockets cleared that gate without any
evidence being brought to bear. An earlier revision of this pipeline produced a shortlist
of four leads on that basis; independent review established that every one was a known
functional site invisible to the gate, in each case because no experimental structure of
the human protein exists. The filters described below resolve known sites through
homologues, and report per gate whether evidence was available at all
(\S\ref{app:coverage}).

\subsection{The ligand-site annotation}
\label{app:qbiolip}

Ligand-contact evidence throughout --- both for the direct gate and as the donor pool for
homology transfer --- comes from a curated structural annotation derived from Q-BioLiP, in which every ligand-binding residue of every deposited structure is
resolved to a canonical UniProt position through per-residue SIFTS. Contacts are
admitted under a fixed policy: a literal 5\,\AA{} heavy-atom contact definition; at least 10
contacting residues per chain; ligands of 5--50 heavy atoms; at least 80\% SIFTS coverage
of the chain with at most a 0.2 mismatch fraction; a single primary accession per chain;
and a UniProt length of 40--1,200 residues.

Two properties of this policy propagate into the results. First, the annotation contains \emph{nonpolymer ligands only} --- no peptide,
nucleic-acid or protein partners --- which is why polymer interfaces are computed
separately from PDB biological assemblies rather than read from the same table. Second,
the 10-residue minimum censors the pocket-size distribution from below. The length
cap is also the reason the direct gate reaches only 2,444 of the 20,092 scanned proteins:
our scan admits chains up to 2,560 residues, so large proteins are annotated in the
reference set only where a shorter construct exists.

\subsection{Donor construction}

Three classes of site were assembled for all organisms, keyed by UniProt accession.
\emph{Ligand contacts} were taken from a cross-species structural annotation of
nonpolymer ligands: 19{,}203 accessions, 322{,}466 ligand pockets and 6{,}122{,}832
contact residues, at a literal 5\,\AA{} contact definition. \emph{Annotated sites} were UniProt
ACT\_SITE and BINDING features, excluding range features spanning more than 30 residues,
which are deletions rather than point annotations. \emph{Interface residues} were
inter-chain contacts computed from PDB biological assemblies and mapped to canonical
numbering through per-residue SIFTS, covering 11{,}999 accessions. Donors carrying at
least one site of any class number 275{,}471, spanning 11{,}231 Pfam families.

\subsection{Transfer procedure}

Each scanned protein was resolved to its Pfam families and, through
\texttt{Pfam-A.clans}, to its clans. Donors sharing a family were ranked first, then
donors sharing only a clan, with ties broken by agreement of the three-level EC number and
then by site count; the top 60 candidates were aligned. Alignment used global
Needleman--Wunsch (parasail, BLOSUM62, gap open 11, extend 1), discarding alignments below
25\% identity over aligned columns, and the 12 highest-identity donors were retained.
Donor site positions were mapped to query numbering through the alignment, and a pocket
was discarded if any lining residue lay within 5\,\AA{} of a transferred position in the
predicted fold.

Two restrictions apply to clan-level transfer. First, a clan is not always a functional
family: CL0015 (MFS) shares a substrate cavity across its members, whereas CL0196 (DSRM)
shares only a small structural motif between otherwise unrelated proteins. Clan-level
transfers therefore require corroboration by at least two independent donors, which
suppresses single spurious alignments without weakening transfer within genuine fold
families. Second, interface transfers are restricted to family level throughout: a binding
cavity is conserved across a clan, but a protein's interaction partners are not. With
these settings the gate reaches 16{,}754 of 20{,}092 proteins.

\subsection{Geometric thresholds}

Thresholds were calibrated against 78{,}370 experimentally determined pocket instances
measured on the same predicted folds. Burial (39.9 protein heavy atoms within 8\,\AA{} of a
lining-residue centroid, excluding the pocket's own atoms) and the upper size bound of 30
residues are the 10th and 95th percentiles of that reference set. The lower size bound is
not a percentile. The reference distribution is censored near ten residues by the
\texttt{minimum\_contact\_residues\_per\_chain} policy of the annotation that produced it,
so it carries no information about smaller pockets and cannot justify a floor in that
range. Lowering the bound without compensation admits low-confidence output --- pockets of
3--5 residues have a median head confidence of 0.29 against 0.82 for those of 16--30 ---
which shows that the size cut had been acting as an implicit confidence filter. We
therefore set the size floor at 9 and impose an explicit confidence floor of 0.70, which
together retain a set of comparable size to the previous thresholds but at higher median
confidence.

Per-residue pLDDT is the mean B-factor over all atoms of a residue in the predicted
structure, and a pocket's pLDDT is the mean over its lining residues. Residues without a
value are excluded from that mean; pockets on proteins for which no per-residue values had
been extracted were previously discarded by the pLDDT gate as missing rather than poor,
affecting 3{,}834 pockets, and were backfilled.

\subsection{Motif-propagated evidence}

Neither direct nor homology-propagated evidence can reach a pocket that is uncharacterized
on its own protein and across its family, yet recurs as a ligand-binding structural motif
in proteins with no detectable evolutionary relationship to it. Each survivor's lining
residues were submitted as a motif to Folddisco against 233{,}054 PDB structures. Matched
target residues were resolved to chain and residue identifiers and intersected with the
ligand-contact annotation of \S\ref{app:qbiolip}; a motif with any matched residue
that is a ligand contact in a deposited structure was discarded. A second rule, flagging
sub-\aa ngstr\"om matches to a target chain of the same family or EC class, fired on a single
pocket and is not load-bearing: most matched target chains carry no annotation to compare
against, so this route is under-powered and the ligand-contact rule carries the filter.

\subsection{Evidence coverage}
\label{app:coverage}

Each gate reports one of three states --- the site is hit, the site is clear, or no
evidence was available --- and each pocket carries the number of gates that had evidence.
Of the 155 survivors, 1 cleared three gates on evidence, 11 cleared two, 54 cleared one
and 89 cleared none. The majority of the output is therefore untested rather than tested
and clear, and should be read as such; homology resolution reduces this fraction (from
87\% to 57\%) but does not eliminate it, because a protein whose Pfam family contains no
annotated member remains unreachable by any transfer.

\subsection{Alternative pocket proposals}

To test whether the head's own decoder limits recall, pocket proposals from fpocket,
P2Rank and Lacuna were generated for all scanned proteins (1{,}797{,}462 candidates) and
scored by the same head. Of those clearing the confidence and size thresholds, 735 on 346
proteins passed the direct and homology-propagated filters. Every one overlapped a pocket the head had
already proposed; none was a site at a location the head had not itself identified, and
none of the proteins for which the head predicted no pocket was rescued. The proposal
stage is therefore not the binding constraint on this pipeline.

\section{Candidate Pockets}
\label{app:high-conf-pockets}

Two pockets are described in the main text. They are not the two highest-scoring
pockets; they are the ones for which a mechanistic hypothesis exists, and they were
selected after the filters rather than by them. Each is reported with the evidence against
it, and with the extent to which conventional cavity detectors agree that the site is a
pocket at all.

\subsection{Agreement with geometric cavity detectors}

fpocket, P2Rank and Lacuna were run natively on the same predicted structures, and each
tool's own score order was used --- not a re-ranking by the head. Table~\ref{tab:detectors}
reports the rank at which each tool first proposes a pocket overlapping ours.

\begin{table}[h]
\centering
\caption{Agreement of geometric cavity detectors with the predicted site. Each tool was
run natively and its own score order used. Entries give the rank of the tool's
best-overlapping proposal, out of the number it made on that protein, with the Jaccard
overlap in parentheses. A dash means no proposal reached a Jaccard of 0.25, so the tool
does not propose this site at any rank. }
\label{tab:detectors}
\begin{tabular}{lcccc}
\toprule
& zPocket & P2Rank & fpocket & Lacuna \\
\midrule
DET1 & 1 & 4 / 17 (0.71) & 39 / 40 (0.70) & 1 / 50 (0.42) \\
COP1 & 1 & --- \phantom{0000}(0.06) & --- \phantom{0000}(0.16) & 24 / 52 (0.42) \\
\bottomrule
\end{tabular}
\end{table}

The two cases differ. For \textbf{DET1} the classical detectors disagree with each other:
P2Rank ranks the site fourth of seventeen and fpocket thirty-ninth of forty, while Lacuna,
itself a learned method, ranks it first. The site is thus found by all three, but a user
relying on fpocket's ordering would never reach it. For \textbf{COP1} no geometric
detector proposes the site at all. P2Rank returns only four proposals on the entire
731-residue protein, of which the highest-scoring is a 40-residue region sharing three
residues with the predicted pocket; the remaining three share none. Only Lacuna proposes a
comparable site, at rank 24 of 52. This is consistent with the site being a shallow peptide
groove, which cavity-geometry methods are least able to score, and it is the case where the
head's proposal is not reproduced by an existing tool.

\subsection{DET1}

DET1 (UniProt Q7L5Y6, 550 residues) is a substrate-adaptor subunit of the
DCX(DET1--COP1) cullin--RING ligase, with RBX1, DDB1, CUL4A and COP1. The predicted
pocket comprises 17 residues drawn from five sequence-distant segments spanning the chain
(57--63, 142--143, 218, 478--479, 517--539), with a mean head confidence of 0.831, burial
of 43.8 heavy atoms and mean pLDDT of 90.1. Because DET1 carries no RING domain and is not
the catalytic subunit, a ligand at this site cannot inhibit the ligase; it can only alter
substrate recognition, which is the architecture exploited by molecular glues. Twelve experimental structures of DET1 are deposited, and across the 16 DET1 chains they
contain, no partner atom approaches within 6\,\AA{} of any lining residue; the nearest partner
contact is 14.9\,\AA{}. The pocket also avoids the E2-binding region and the DDB1-anchoring
segment.

\paragraph{Limitations.} The site is not a pre-formed cavity. Across all 16 experimental
copies the lining is solvent-exposed, with a mean relative SASA of 0.295 and no fully
buried lining residue, and fpocket ranks the overlapping groove in the bottom half of its
own list. This measurement requires care: computed on holo structures with the ligand
deleted, the same statistic rejects apo cereblon, which scores worse than DET1 --- every
validated site appears enclosed once a ligand has opened it. Measured against genuinely
apo references the result is that DET1's site is not pre-formed, not that it is
unligandable, and the lining is mildly more mobile than the protein average (1.31\,\AA{}
side-chain spread against 1.09\,\AA{}).

\subsection{COP1}

COP1 (Q8NHY2, 731 residues) is the substrate receptor of the same ligase complex as DET1,
degrading JUN, p53 and the ETV1/4/5 transcription factors. The predicted pocket comprises 15 residues, all 15 inside annotated WD40 repeats and
spanning five of the seven blades --- the top face of the $\beta$-propeller. Superposition
on the COP1--TRIB1 complex shows that 14 of the 15 are within 4.5\,\AA{} of the bound degron
peptide, and that the three-residue Val--Pro anchor of that peptide engages ten COP1
residues, all ten of which are in the predicted pocket. The site is therefore not merely
the degron surface but specifically its hot-spot subsite, which is the configuration that
makes a protein--protein interface tractable to small molecules.

\paragraph{Limitations.} Occupancy blocks substrate recognition, so a ligand is first an inhibitor, and a
cereblon-style glue is not available at this pocket. Direction of effect is substrate-dependent and cannot be
separated chemically, since p53, C/EBP$\alpha$ and ETV4/5 all dock through the same slot.
The site has been attacked: an affinity-selection campaign reported six COP1-binding
peptides at potencies comparable to the native motif, and no small molecule is known;
ChEMBL contains no COP1 target record at all. The experimental literature describes the
VP pocket as shallow.

\end{document}

%% file: figures/tables.tex
\newcommand{\tabdatasetbody}{%
Train & 11{,}016 (9{,}362) & 14{,}331 (11{,}850) & 4{,}628{,}164 (4{,}037{,}795) & 6.32\% (6.18\%) \\
Validation & 1{,}553 (1{,}266) & 2{,}165 (1{,}508) & 724{,}783 (585{,}943) & 5.75\% (5.12\%) \\
Test & 1{,}566 (1{,}283) & 2{,}301 (1{,}577) & 766{,}340 (629{,}524) & 5.73\% (4.85\%) \\
\midrule
Effective test set & 1{,}275 & 1{,}562 & 625{,}883 & 4.86\% \\
\midrule
\textbf{Total} & \textbf{14{,}135 (11{,}911)} & \textbf{18{,}797 (14{,}935)} & \textbf{6{,}119{,}287 (5{,}253{,}262)} & \textbf{6.18\% (5.90\%)} \\
}
\newcommand{\ntargets}{1{,}275}
\newcommand{\npockets}{1{,}562}

\newcommand{\testchains}{1{,}279}
\newcommand{\testpockets}{1{,}573}

\newcommand{\testposrate}{4.86}
\newcommand{\testsubsumed}{4}

\newcommand{\ntestpocketstage}{1{,}577}
\newcommand{\ndroppedpockets}{11}
\newcommand{\toppocketpct}{71.6}